\documentclass[10pt,twocolumn,letterpaper]{article}

\usepackage[pagenumbers]{iccv} 

\definecolor{iccvblue}{rgb}{0.21,0.49,0.74}
\usepackage[pagebackref,breaklinks,colorlinks,allcolors=iccvblue]{hyperref}

\newcommand{\bftable}{\fontseries{b}\selectfont}
\usepackage{multirow}
\usepackage{xspace}
\usepackage{tcolorbox}
\usepackage{colortbl}       
\usepackage{makecell}
\usepackage[accsupp]{axessibility}

\definecolor{featherbrown}{HTML}{8C564B}
\definecolor{featherbackgroundbrown}{HTML}{F1E1DB}
\definecolor{fastvgreen}{HTML}{3B7D23}
\definecolor{fastvbackgroundgreen}{HTML}{EDF8E7}
\definecolor{localizationpurple}{HTML}{78206E}
\definecolor{localizationbackgroundpurple}{HTML}{F8E7F6}

\definecolor{myred}{rgb}{0.6274, 0, 0}
\definecolor{myblue}{rgb}{0.416, 0.612, 0.969}
\definecolor{mygreen}{rgb}{0.267, 0.596, 0.361}

\newcommand\feather{\textcolor{featherbrown}{\texttt{\textbf{FEATHER}}}\xspace}

\newcommand\featherlong{\textbf{F}ast and \textbf{E}ffective \textbf{A}cceleration wi\textbf{TH} \textbf{E}nsemble c\textbf{R}iteria\xspace}

\newcommand{\rotatecol}[1]{\rotatebox{90}{#1}}
\newcommand{\colpad}{\hspace{6pt}}
\newcommand{\colorcell}{\cellcolor{iccvblue!10}}
\newcommand{\colorcellgray}{\cellcolor{gray!5}}

\def\paperID{10505} 
\def\confName{ICCV}
\def\confYear{2025}

\title{Feather the Throttle: Revisiting Visual Token Pruning for \\ Vision-Language Model Acceleration}

\author{
Mark Endo, Xiaohan Wang, Serena Yeung-Levy\\
Stanford University\\
{\tt\small \{markendo,xhanwang,syyeung\}@stanford.edu}
}

\begin{document}
\maketitle
\begin{abstract}

Recent works on accelerating Vision-Language Models achieve strong performance across a variety of vision-language tasks despite highly compressing visual information. In this work, we examine the popular acceleration approach of early pruning of visual tokens inside the language model. Surprisingly, we find that while strong performance is maintained across many tasks, it exhibits drastically different behavior for a subset of vision-centric tasks such as localization. Upon further investigation, we uncover a core issue with the acceleration approach where most tokens towards the top of the image are pruned away. Yet, on many benchmarks aiming to evaluate vision-centric capabilities, strong performance persists with the flawed pruning strategy, highlighting these benchmarks' limited ability to assess fine-grained visual capabilities. Based on these findings, we propose \feather (\featherlong), a straightforward approach that resolves the discovered early-layer pruning issue and further enhances the preservation of relevant tokens via multistage pruning with early uniform sampling to ensure broad image coverage. With comparable computational savings, we find that \feather achieves more than $\mathbf{5\times}$ performance improvement on the vision-centric localization benchmarks compared to the original acceleration approach.

\end{abstract}    
\section{Introduction}
\label{sec:intro}

\begin{figure}[t]
  \centering
   \includegraphics[width=1\linewidth]{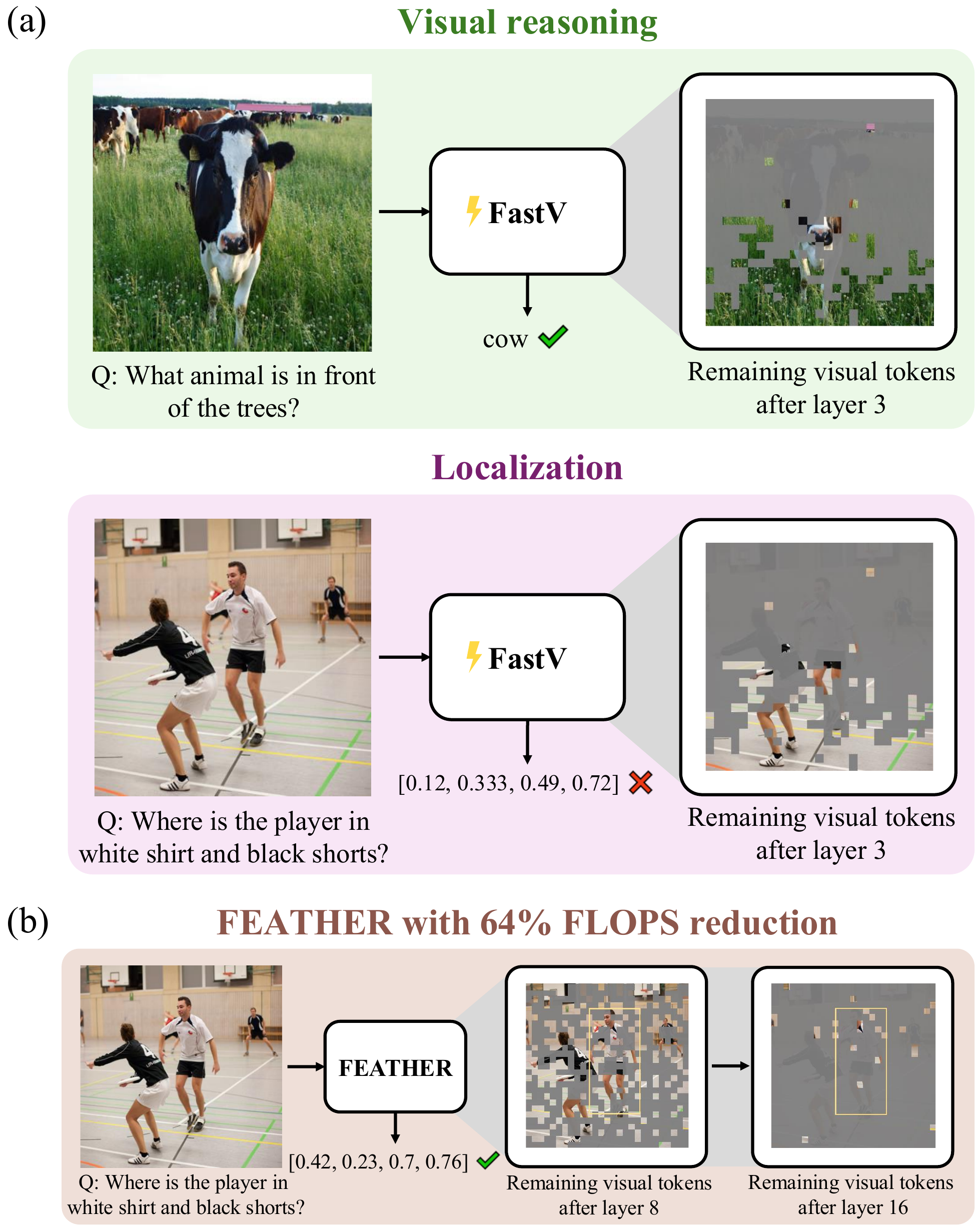}
   \caption{\textbf{(a)} Although FastV prunes most visual tokens from the upper portion of the image, the approach still displays strong performance on a variety of evaluated vision-language tasks except for a small subset of vision-centric tasks like localization.
   \textbf{(b)} Based on our findings, we propose \feather (\featherlong), a straightforward approach that resolves the existing issue of selecting bottom tokens, additionally maintains uniformly sampled tokens to ensure good coverage over the whole image, and prunes in two stages.}
   \label{fig:pull_fig}
\end{figure}

The exploration of Vision-Language Models (VLMs) is a critical area of computer vision and natural language processing research, centered on combining large language models (LLMs) with visual encoders to enable multi-modal perception, reasoning, and understanding capabilities. While earlier works explored sophisticated schemes for conditioning language models with visual information \cite{alayrac2022flamingo, li2022blip, li2023blip2}, more recently the space has shifted to predominately using the simplistic approach of taking patch features from pre-trained visual encoders and projecting them to the input space of the language model with a light-weight adapter \cite{liu2024llava, liu2024llava1.5, bai2023qwenvl, wang2024qwen2vl}. Using image patches as tokens, however, comes with the drawback of being computationally inefficient. To achieve fine-grained resolution, the image is divided into many patches. This large number of patches significantly increases computational demands due to the quadratic complexity of the attention operation in Transformers. As a result, many recent works have focused on accelerating these methods by compressing visual information, demonstrating that heavy compression can still maintain strong performance across a wide variety of tasks \cite{chen2024fastv, xing2024pyramiddrop, shang2024prumerge, arif2024hired, cai2024matryoshka}. For instance, FastV \cite{chen2024fastv} prunes 50\% of visual tokens after the shallow layers of the Language Model (LLM) while not compromising in performance across a range of image and video understanding tasks \cite{chen2024fastv}. Another work reveals that for a fixed computational budget on visual reasoning tasks, optimal performance is achieved by using the largest LLM possible while sacrificing visual information, often reducing the visual token count to a single token \cite{li2024onetoken}. With such compressed visual information, it is still unclear how these methods achieve high performance on tasks assessing vision capabilities such as visual reasoning and understanding, and whether there are more demanding visual tasks where these methods fail.

The motivation of our study is to get a better understanding of the vision capabilities of accelerated VLMs given that they leverage highly compressed visual information,  focusing specifically on FastV approach. Evaluating across a wide range of vision-language tasks, we find that while the approach maintains strong performance across many tasks, there is a substantial decrease in performance for TextVQA and a severe decrease for localization tasks. When analyzing the approach's poor performance, we uncover a \textit{fundamental} issue with the pruning criteria in early layers where tokens towards the top of the image are disproportionally removed due to the long-term decay property of RoPE. Since this issue is not task-specific, we examine how performance is maintained across most evaluated tasks and find that even when tokens are entirely discarded using the flawed criteria, performance remains largely unchanged. This highlights how many benchmarks require minimal visual grounding. Our finding underscores a significant challenge in the field of multimodal learning, not only for measuring the effectiveness of VLM acceleration methods but also for benchmarking the visual capabilities of VLMs as a whole.

Given the discovered limitation of the studied approach, we next explore whether the attention-based criteria can be modified to enable robust visual token pruning, leading to our final method, \feather (\featherlong).  Specifically, we propose a straightforward modification to the attention-based criteria by removing RoPE, demonstrating that this resolves the positional bias issue and greatly improves performance. Additionally, we incorporate uniform sampling in early token pruning to ensure adequate coverage of all image regions and apply more extensive pruning at a later layer, when the attention-based criteria better discerns important tokens. This strategy is analogous to how a racecar driver \textit{feathers the throttle} by gradually pressing the accelerator at the beginning of a turn to maintain grip and then accelerating more aggressively once the car is past the apex. We show that \feather results in substantial performance gains compared to the original acceleration approach, improving localization performance more than $\mathbf{5\times}$ with comparable computational savings. Strikingly, we find that our approach achieves this performance improvement while only retaining 3.3\% of visual tokens for the second half of LLM layers. Overall, our work demonstrates that while visual compression can maintain strong performance even on challenging vision-centric tasks, its effectiveness depends on a well-designed strategy, which is currently difficult to assess due to many vision-language benchmarks not thoroughly evaluating vision capabilities.

In summary, our contributions include:

\begin{itemize}
    \setlength{\leftskip}{5mm}
    \item \textbf{Revealing inconsistent effect of early pruning:} Our evaluations across 12 benchmarks show that early visual token pruning minimally affects performance on most tasks, but causes a substantial dropoff for TextVQA and a drastic decline for localization tasks.
    \item \textbf{Analyzing poor performance:} We uncover that the attention-based criteria for selecting important tokens exhibits a bias towards bottom image tokens in early layers due to the long-term decay property of RoPE.
    \item \textbf{Identifying benchmark limitations in assessing token pruning:} We find that performance remains strong even with suboptimal pruning criteria and removal of pruned tokens before transferring information in the LLM, demonstrating how many benchmarks fail to assess fine-grained visual capabilities.
    \item \textbf{Improving attention-based criteria:} With the straightforward modification of removing RoPE, we demonstrate that attention can be used as an effective criteria for selecting tokens and performance can be further improved with simple strategies like incorporating uniform sampling and pruning more extensively at a later layer, when the criteria better discerns important tokens.
\end{itemize}
\section{Related Work}
\label{sec:related_work}

Recent efforts to accelerate VLMs can be broadly divided into two main categories: compressing visual information before it enters the LLM, and compressing visual information within the LLM itself. In the first category, Chat-UniVi \cite{jin2024chatunivi} dynamically merges visual tokens with similar semantic meanings. PruMerge \cite{shang2024prumerge}, FasterVLM \cite{zhang2024fastervlm}, and VisionZip \cite{yang2025visionzip} select important tokens based on ViT attention ([CLS] or average attention received). PruMerge then merges important tokens with unselected ones while VisionZip merges unselected tokens together. For the LLaVA-NeXT \cite{liu2024llavanext} approach of partitioning an image into sub-images where inefficiency is an even bigger problem, HiRED \cite{arif2024hired} selects tokens with top feature important on each sub-image with an allocated budget. Other methods argue that the input image alone does not include enough information to select important patches and thus use the textual input to recover visually meaningful tokens \cite{chen2024recoverablecompression, yu2024balancingperformanceefficiency}. Alternatively, some approaches reduce visual information within the ViT by merging or pruning tokens \cite{chai2024auroracap, liu2024revisitingtokenpruning, bolya2023tokenmerging}.

For the second category of approaches, where visual information is compressed within the LLM itself, LOOK-M reduces the multimodal KV cache size \cite{wan2024look}. In our study, we focus on the popular FastV approach \cite{chen2024fastv}. This work identifies that attention over image tokens is sparse in deeper layers of the LLM. Based on this observation, they propose to prune away unimportant vision tokens after the shallow layers of the LLM, achieving a 45\% reduction in FLOPS with nearly no performance loss. Works since have proposed alterations to this setup such as adaptively determining the number of pruned tokens instead of using a fixed ratio \cite{he2024zipvl} or pruning in multiple stages \cite{xing2024pyramiddrop}.

\section{Revising Visual Token Pruning}
\label{sec:hidden_cost_of_vlm_inference_acceleration}

In this section, we aim to get a better understanding of the vision capabilities of the VLM acceleration approach of pruning visual tokens after shallow LLM layers. After outlining preliminaries (\S\ref{sec:preliminaries}), we take a closer look at how the approach performs across a broad range of tasks. Upon inspection, we discover that while heavily pruning visual tokens after the shallow LLM layers has little effect on performance across a variety of tasks, this approach fails decisively on more vision-centric tasks, particularly localization (\S\ref{sec:bad_at_localization}). Next, we examine why this method struggles with localization, uncovering that the poor performance is due to the ineffectiveness of the pruning criteria to select important tokens when applied at an early layer (\S\ref{sec:fastv_criteria}). As this defect is not specific to localization, we explore what can be attributed to the method's high performance on numerous other tasks and find that these tasks require minimal visual grounding (\S\ref{sec:vlm_inference_on_other_tasks}).

\begin{figure*}[th]
  \centering
   \includegraphics[width=1\linewidth]{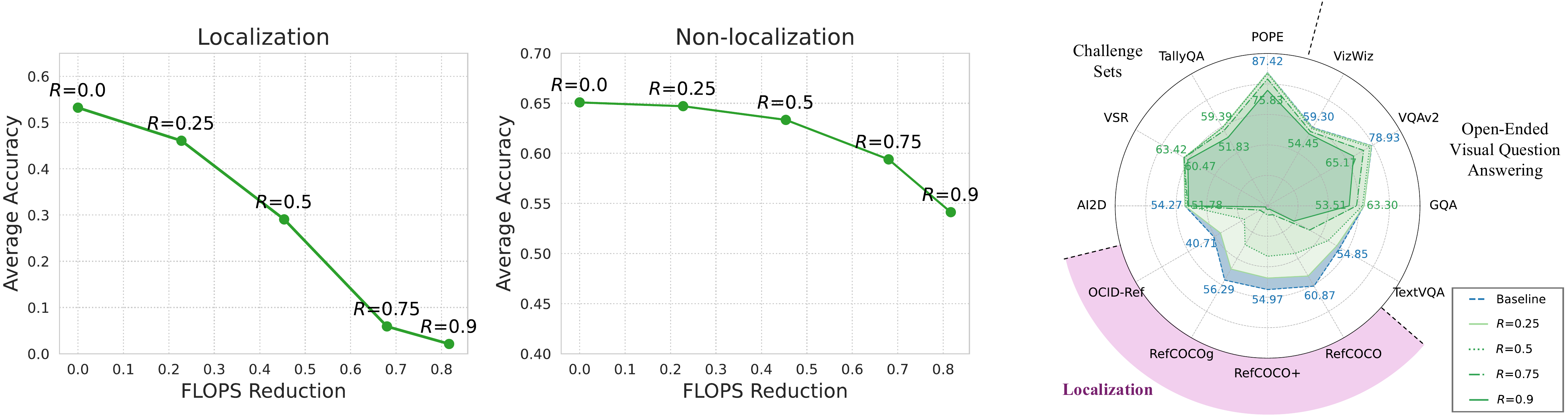}
   \caption{Contrasting the difference in performance dropoff on the challenging vision-centric localization task (\textbf{Left}) versus the other evaluated tasks (\textbf{Middle}) when pruning visual tokens after the shallow LLM layers. Whereas performance decrease is minimal for most tasks, localization exhibits roughly a linear decrease to zero as the ratio of pruned tokens increases. \textbf{Right:} Per-task performance breakdown across various setups of pruning ratios. Using $\textit{K} = 3$ for all pruning setups.}
   \label{fig:hidden_cost_results}
\end{figure*}

\subsection{Preliminaries}
\label{sec:preliminaries}

Before our analyses, we provide a background on the explored adapter-style VLM with visual token pruning, the evaluated benchmarks, and experimental settings.

\textbf{VLM and token pruning.} Formally, an adapter-style VLM takes as input an image $x_{\text{img}}$ and text prompt tokens $x_{\text{prompt}}$. A pre-trained vision backbone $f$ first encodes visual features $z_{\text{img}} = f(x_{\text{img}}) \in \mathbb{R} ^{n \times d_{\text{vision}}}$ where $n$ is the number of image patches and $d_{\text{vision}}$ is the dimensionality of the vision encoder. Next, an adapter $p$ (either a simple linear layer or MLP) projects the vision features to embeddings $h_{\text{img}} = p(z_{\text{img}})\in \mathbb{R} ^{n \times d_{\text{text}}}$ where $d_{\text{text}}$ is the dimensionality of the LLM. Lastly, $h_{\text{img}}$ is concatenated with text prompt embeddings $h_{\text{prompt}} = \text{embed}(x_{\text{prompt}})$ and passed into the language model LM to generate the output text $y = \text{LM}([h_{\text{img}};h_{\text{prompt}}])$.

In this work, we study the inference acceleration of VLMs where visual tokens are pruned within the attention mechanism of LM. We focus on the FastV \cite{chen2024fastv} approach, where after layer $K$ in LM, $R$\% of visual tokens are pruned away based on a ranking function $g_{\phi}$ which ranks tokens based on criteria $\phi$. In practice, the attention score received from the last text token is used as the criteria, referred to as $\phi_{\text{original}}$. Note that the positional information is preserved when performing the pruning.

With this approach, we measure the acceleration using the theoretical FLOPS reduction ratio related to the image tokens. For one Transformer layer, the total FLOPS is estimated as $C = 4nd^2 + 2n^2d + 2nd_m$ where $d = d_{\text{text}}$ and $m$ is the intermediate size of FNN. Given that the Transfoerm has $T$ layers in total and after layer $K$, we maintain $\hat n= (1 - R\%) * n$ visual tokens, we calculate the FLOPS reduction as 
\begin{small}
\begin{equation}
    1 - \frac{K * C  + (T - K) * (4\hat nd^2 + 2\hat n^2d + 2\hat ndm)}{C}.
\end{equation}
\end{small}

\textbf{Benchmarking.} We evaluate the accelerated VLMs on a large suite of benchmarks from \cite{karamcheti2024prismatic} which includes evaluations spanning the areas of localization, open-ended visual question answering, and challenge sets. For all benchmarks, we follow the same evaluation protocol as \cite{karamcheti2024prismatic}.

\textit{Localization.} We evaluate localization performance using the RefCOCO, RefCOCO+, RefCOCOg \cite{kazemzadeh2014referitgame, yu2016refcoco} and OCID-Ref \cite{wang2021-ocidref} datasets. RefCOCO contains short, spatially grounded descriptions, while RefCOCO+ focuses on appearance-based descriptions. RefCOCOg, on the other hand, features longer and more detailed descriptions. OCID-Ref is a robotics dataset that tests generalization to out-of-distribution scenarios, particularly for object localization in cluttered environments.

\textit{Open-Ended Visual Question Answering.} We evaluate general visual reasoning using the VizWiz \cite{bigham2010vizwiz} and VQAv2 \cite{goyal2017vqav2} datasets, spatial reasoning using the GQA \cite{hudson2019gqa} dataset, and reasoning around text using the TextVQA \cite{singh2019textvqa} dataset. For TextVQA, following \cite{karamcheti2024prismatic}, we exclude OCR-system parsed input tokens to better evaluate the effect of pruning on visual capabilities.

\textit{Challenge Sets.} We additionally evaluate on the VSR \cite{liu2023vsr}, TallyQA \cite{acharya2019tallyqa}, POPE \cite{li2023pope}, and AI2D \cite{kembhavi2016ai2d} closed-set prediction datasets. VSR includes binary spatial relationship questions, TallyQA consists of counting questions, POPE probes hallucination, and AI2D contains multiple-choice questions referring to scientific diagrams and charts.

\textbf{Experimental settings.} In our experiments, we utilize a VLM with SigLIP ViT-SO400M \cite{zhai2023siglip} as $f$, a one-layer MLP with GELU activation as $p$, and Llama 2 7B \cite{touvron2023llama2} as LM. The model was trained on the multimodal instruction tuning dataset presented in \cite{liu2024llava1.5} in a single-stage with $f$ frozen.

\subsection{Impact of early visual token pruning varies drastically by task}
\label{sec:bad_at_localization}

We examine the effect of pruning visual tokens after the shallow LLM layers ($K = 3$) across a variety of tasks. We compare various pruning ratios ($R \in \{0.25, 0.5, 0.75, 0.9\}$) and the baseline non-pruned model in Figure~\ref{fig:hidden_cost_results}.

For the majority of evaluated tasks, heavily reducing the number of visual tokens results in minimal performance decrease (3.5\% for AI2D, 0.1\% for VSR, 3.7\% for TallyQA, 4.8\% for POPE, 5.1\% for VizWiz, 7.9\% for VQAv2 and 7.7\% for GQA when dropping 75\% of tokens).
\textbf{However, for a small subset of tasks--particularly TextVQA, and more severely, localization--performance drops substantially}, with decreases of 42.0\%, 88.9\%, 88.9\%, 91.0\%, and 86.0\% for TextVQA, RefCOCO, RefCOCO+, RefCOCOg, and OCID-Ref, respectively.
Investigating further, Figure~\ref{fig:hidden_cost_results} shows that localization performance declines roughly linearly to zero as the pruning ratio progresses from 0 to 1, a pattern that strongly contrasts with most other tasks, where performance remains largely unaffected. While prior works report a much smaller decline in TextVQA performance \cite{xing2024pyramiddrop, chen2024llavolta}, they incorporate OCR-parsed input tokens into prompts, reducing the reliance on visual information.

\noindent \textbf{\textit{Discussion}}. Although localization is undoubtedly challenging for a model that discards visual tokens, as it requires fine-grained visual grounding and precise boundary identification, it is still surprising that (1) localization performance declines to the extent it does even when many tokens are maintained  (e.g., dropping 50\% of tokens reduces average performance by 45\%), and (2) most non-localization benchmarks, except for TextVQA,  exhibit a drastically different pattern despite these benchmarks aiming to evaluate visual grounding abilities, such as counting for TallyQA, spatial reasoning for GQA, and chart understanding for AI2D. In the following sections, we seek to explain these unexpected findings and understand their implications.

\begin{figure*}[th]
  \centering
   \includegraphics[width=1\linewidth]{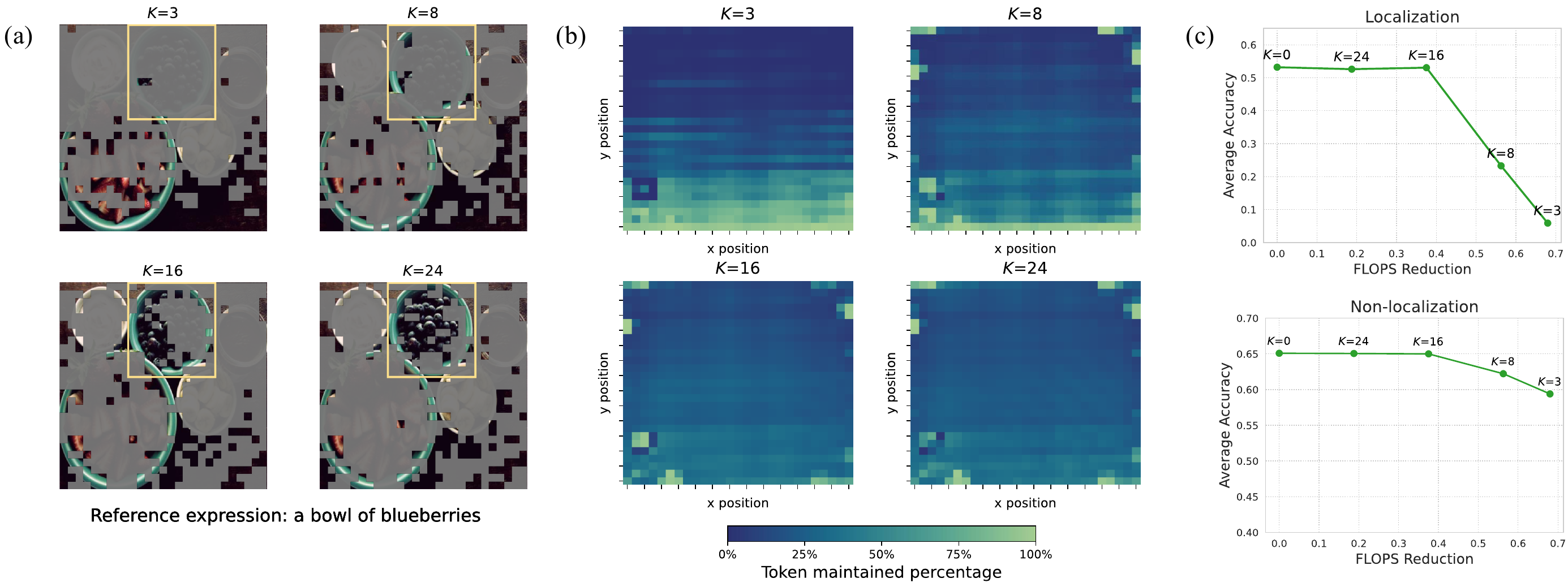}
   \caption{\textbf{(a)} Example demonstrating that when pruning in early layers, selected tokens are concentrated in the bottom of the image, whereas in later layers the selected tokens more precisely cover the region of the image important for answering the question. \textbf{(b)} Heatmap illustrating that averaged across all benchmark examples, as the pruning layer increases, the selection of bottom visual tokens by the criteria is reduced. \textbf{(c)} Visualizing the effect of the pruning layer on performance for both localization and non-localization tasks. We find that pruning after layer 16 or later results in little performance dropoff, whereas pruning earlier results in a performance decrease, particularly for localization. Using $R = 0.75$ for all setups.}
   \label{fig:fastv_ineffective_criteria}
\end{figure*}

\subsection{Interpreting poor vision-centric task performance}
\label{sec:fastv_criteria}

We next investigate why pruning visual tokens after the shallow LLM layers has such an adverse effect on localization performance. Intuitively, even localization should not require all image tokens, as there are background tokens irrelevant to identifying specific objects that can be pruned. Thus, it is surprising that performance begins to degrade even with minimal token pruning. This suggests that the pruning criteria, intended to retain important tokens, fails to effectively distinguish which tokens contain necessary visual information.

To assess the efficacy of the pruning criteria, we first examine the distribution of retained tokens across all benchmark examples. As shown in Figure~\ref{fig:fastv_ineffective_criteria}(b), we find that \textbf{pruning visual tokens after the shallow LLM layers ($K = 3$) retains tokens concentrated at the bottom of the image}. For example, when $R=0.75$, the average sampled token position across all datasets is situated 80.7\% of the way down the image. A Chi-Square test confirms the non-uniformity of selected tokens' y-positions ($p$-value $<$ 0.05). We also examine the criteria behavior when pruning after later layers ($K \in (8, 16, 24)$) and observe that the bias of selecting bottom image tokens is reduced.

In addition, we observe that as the pruning layer increases, not only is the positional bias reduced, but the criteria also begins to correctly select tokens relevant to the text instruction. For example, as shown in Figure~\ref{fig:fastv_ineffective_criteria}(a), pruning after the later layers results in a distinct concentration of selected tokens around the area corresponding to the reference expression for localization. To quantify this change in criteria effectiveness across layers, we measure the performance when varying the pruning layer ($K \in \{3, 8, 16, 24\}$). As shown in Figure~\ref{fig:fastv_ineffective_criteria}(c), performance remains stable when pruning is performed at deeper layers ($K=16$ and $K=24$), whereas pruning at earlier layers ($K=3$ and $K=8$) leads to performance degradation, particularly for localization.

\noindent \textbf{\textit{Discussion}}. While previous research attributes the poor performance of early token pruning to the low redundancy of image tokens in shallow layers \cite{xing2024pyramiddrop}, we reveal a critical insight: \textbf{the variable performance across pruning layers is linked to the effectiveness of selecting important tokens, as early token pruning leads to suboptimal token selection biased toward bottom tokens.} To understand why the criteria disproportionately retains bottom image tokens, we must consider the LM architecture. Specifically, token selection is based on the attention received from the last text token. Since the LM uses RoPE to encode positional information \cite{su2024roformer}, it exhibits a long-term decay property. While this property is well-suited for language modeling, we uncover that it poses an issue for selecting visual tokens, where tokens appearing later in the flattened, raster-scan order have higher attention scores purely due to their position. In language modeling, prior work finds that there is greater emphasis on shorter-distance information in shallow layers \cite{hong2024token}. We predict that the visually-conditioned LM exhibits the same behavior, explaining why the positional bias is most pronounced when pruning after shallow layers.

\begin{figure}[ht]
  \centering
   \includegraphics[width=1\linewidth]{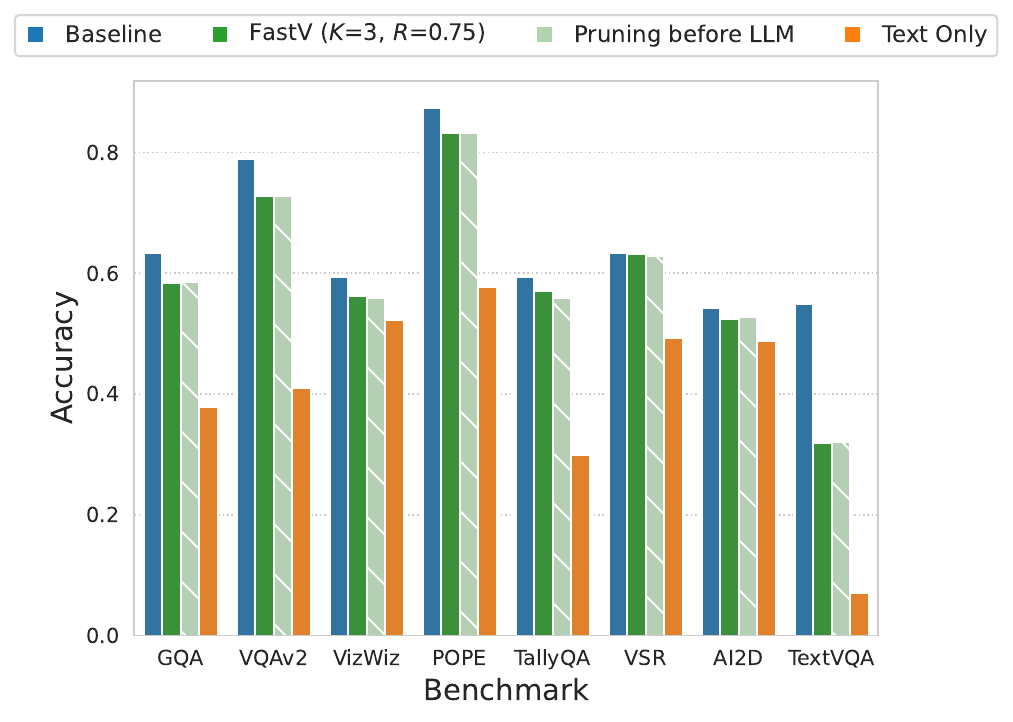}
   \caption{Assessing whether the strong performance of early visual token pruning for many tasks can be attributed to visual information transfer before pruning or benchmarks' lack of assessing fine-grained visual capabilities. We observe that for all evaluated tasks, allowing information transfer before pruning (shown in green) does not result in substantial performance improvement over the setup without visual information transfer (shown in light green), highlighting a limitation of many benchmarks.}
   \label{fig:other_task_performance}
\end{figure}

\subsection{Explaining strong performance on other tasks}
\label{sec:vlm_inference_on_other_tasks}

Given that early pruning of visual tokens leads to a suboptimal tokens selection where remaining tokens are concentrated towards the bottom of the image, a natural question arises: \textit{how does the approach still maintain high performance across a diverse range of tasks, including those aiming to evaluate visual grounding abilities?} In this section, we test the following contrasting explanations:

\begin{enumerate}
    \item Important visual information from top image tokens is transferred to later tokens before pruning with the unidirectional attention of the LLM.
    \item Many questions can still be inferred with access to only the suboptimal set of visual tokens primarily from the bottom of the image.
\end{enumerate}

To evaluate these hypotheses across tasks, we save the tokens retained by the studied pruning approach and, in a subsequent run, remove all other tokens before they reach the LLM, ensuring that no visual information from the pruned tokens is retained. As an additional comparison, we also evaluate a text-only model in which all visual tokens are removed before entering the LLM. As shown in Figure~\ref{fig:other_task_performance}, we find that across all evaluated tasks, \textbf{removing pruned tokens before the LLM results in little to no performance decrease compared to the method that allows information transfer in shallow layers prior to pruning.}

\noindent \textbf{\textit{Discussion}}. This finding indicates that the high performance of early visual token pruning on many tasks does not stem from the effectiveness of visual information transfer in early layers but rather signifies that many benchmarks do not demand a detailed understanding of visual information to answer questions accurately. Note that for the majority of these benchmarks (except VizWiz and AI2D), performance substantially declines when not including any visual information, signifying that simply comparing vision-enabled and disabled setups is insufficient to determine a benchmark's ability to assess visual grounding.

\section{Improving Visual Token Pruning}
\label{sec:feather}

Based on our findings from \S\ref{sec:hidden_cost_of_vlm_inference_acceleration}, we now seek to answer: \textit{can attention serve as an effective criteria for token pruning even in early layers?}
To this end, we propose a straightforward solution to removing the positional bias of the attention-based criteria and demonstrate that our modifications enable effective pruning even when applied after early LLM layers (\S\ref{sec:improving_pruning_criteria}).
Guided by our insights, we present our final approach, \feather (\featherlong) (\S\ref{sec:distilling_insights}), and showcase its impressive visual capabilities while offering high computational efficiency (\S\ref{sec:feather_results}).

\subsection{Assessing potential of attention-based criteria}
\label{sec:improving_pruning_criteria}

An attention-based criteria offers the potential to naturally identify and retain only the most relevant visual tokens for answering a question. However, as shown in \S\ref{sec:hidden_cost_of_vlm_inference_acceleration}, positional bias strongly influences token selection, limiting its effectiveness. Thus, we aim to resolve the positional bias issue and explore the true promise of attention-based criteria.  We compare against non-attention-based approaches that incorporate an explicit inductive bias toward reducing redundancy. Below, we outline the explored criteria.

$\phi_{\textbf{\textit{-R}}}$: Given the long-term decay property of RoPE, we propose the following straightforward adjustment to the pruning criteria in order to mitigate this bias and more effectively select tokens relevant to the text instruction. Namely, we still compute the attention score received from the last text token, except we do not apply RoPE to the attention mechanism, thereby removing the long-term decay effect. Note that this lightweight calculation is only done once per pruning, and all other state computations are compatible with FlashAttention \cite{dao2022flashattention}.

$\phi_{\text{uniform}}$: In this simple criteria, we uniformly sample visual tokens in the image, using a set stride. This criteria ensures that there is good image coverage, but does so by sacrificing the ability to have more densely captured visual information for a particular image region. To use a similar computational saving to the attention-based criteria when $R=0.75$, we use a stride of two, resulting in 196 selected tokens. 

$\phi_{\text{KNN}}$: Inspired by dynamic visual tokenization \cite{jin2024chatunivi}, we select tokens based on local density. For each visual token $z_{img}[i]$ where $i \in \{0, 1, ..., n-1\}$, we compute its local density $\rho_i$ using K-nearest neighbors and calculate distance index $\delta_i$ as:
\begin{small}
\begin{equation}
\delta_i =
\begin{cases}
\underset{j: \rho_j > \rho_i}{\min} \lVert z_{img}[i] - z_{img}[j] \rVert ^2, & \text{if } \exists j \text{ s.t. } \rho_j > \rho_i. \\
\underset{\quad j \quad}{\max} \lVert z_{img}[i] - z_{img}[j] \rVert ^2, & \text{otherwise.}
\end{cases}
\end{equation}
\end{small}

$\delta_i$ represents how far away $z_{img}[i]$ is from other high-density tokens. We use $\rho_i * \delta_i$ as the token importance score for the criteria and select 196 tokens.

$\phi_{\textbf{\textit{-R}}} + \phi_{\text{uniform}}$: Since $\phi_{\textbf{\textit{-R}}}$ focuses on selecting important tokens and $\phi_{\text{uniform}}$ aims to reduce redundancy, we propose an ensemble of these two to leverage their respective strengths. Specifically, we apply $\phi_{\textbf{\textit{-R}}}$ while incorporating a small number of uniformly sampled tokens using a stride of three.

\textbf{Improvement of removing RoPE:} In Table~\ref{tab:fastv_alternatives}, we report results evaluating the effectiveness of these criteria. Comparing $\phi_{\textbf{\textit{-R}}}$ with $\phi_{\text{original}}$, we find that by removing RoPE, at $K=3$ there is a 183\% average improvement on localization tasks and at $K=8$, there is a 17\% average improvement. These performance improvements demonstrate that once the impact of token position on attention is removed, attention score can more effectively be used for the criteria when applied after early LLM layers. Note that the narrowing gap in performance between $\phi_{\textbf{\textit{-R}}}$ and $\phi_{\text{original}}$ from $K=3$ to $K=8$ is likely because $\phi_{\text{original}}$ has less of a bias towards selecting bottom image tokens as the pruning layer increases.

\begin{table*}[t]
  \centering
 \begin{small}
  \begin{tabular}{cr@{\colpad}c@{\colpad}|@{\colpad}c@{\colpad}c@{\colpad}c@{\colpad}c@{\colpad}c@{\colpad}|@{\colpad}c@{\colpad}c@{\colpad}c@{\colpad}c@{\colpad}c@{\colpad}|@{\colpad}c@{\colpad}c@{\colpad}c@{\colpad}c@{\colpad}c@{\colpad}}
  \multicolumn{3}{c}{} & \multicolumn{5}{c}{Localization} & \multicolumn{5}{c}{Open-Ended VQA} & \multicolumn{5}{c}{Challenge Sets} \\
  \shortstack{Pruning\\Layer} & Criteria & \rotatecol{FLOPS Red \vphantom{X}} & \rotatecol{Avg} & \rotatecol{OCID-Ref} & \rotatecol{RefCOCOg} & \rotatecol{RefCOCO+} & \rotatecol{RefCOCO} & \rotatecol{Avg} & \rotatecol{TextVQA} & \rotatecol{GQA} & \rotatecol{VQAv2} & \rotatecol{VizWiz} & \rotatecol{Avg}  & \rotatecol{POPE} & \rotatecol{TallyQA} & \rotatecol{VSR} & \rotatecol{AI2D} \\
  \midrule
  & \colorcellgray \textit{Attention-based} & \colorcellgray & \colorcellgray & \colorcellgray & \colorcellgray & \colorcellgray & \colorcellgray & \colorcellgray & \colorcellgray & \colorcellgray & \colorcellgray & \colorcellgray & \colorcellgray & \colorcellgray & \colorcellgray & \colorcellgray & \colorcellgray \\
  \multirow{8}{*}{$K = 3$} & $\phi_{\text{original}}$ & 68\% & 5.9 & 5.7 & 5.1 & 6.1 & 6.7 & 54.8 & 31.8 & 58.4 & 72.7 & 56.3 & 64.0 & 83.2 & 57.1 & \bftable 63.3 & 52.4 \\
  & \colorcell $\phi_{\textbf{\textit{-R}}}$ (Ours) & \colorcell 68\% & \colorcell 16.7 & \underline{\colorcell 22.9} & \colorcell 15.1 & \colorcell 13.3 & \colorcell 15.3 & \underline{\colorcell 59.0} & \underline{\colorcell 41.6} & \colorcell 61.2 & \underline{\colorcell 76.0} & \underline{\colorcell 57.3} & \underline{\colorcell 64.7} & \underline{\colorcell 85.2} & \underline{\colorcell 58.2} & \colorcell 62.2 & \underline{\colorcell 53.2} \\
  & \scriptsize \textcolor{mygreen}{$\Delta$} & & \scriptsize \bftable \textcolor{mygreen}{+10.7} & \scriptsize \bftable \textcolor{mygreen}{+17.2} & \scriptsize \bftable \textcolor{mygreen}{+10.0} & \scriptsize \bftable \textcolor{mygreen}{+7.3} & \scriptsize \bftable \textcolor{mygreen}{+8.6} & \scriptsize \bftable \textcolor{mygreen}{+4.2} & \scriptsize \bftable \textcolor{mygreen}{+9.7} & \scriptsize \bftable \textcolor{mygreen}{+2.8} & \scriptsize \bftable \textcolor{mygreen}{+3.2} & \scriptsize \bftable \textcolor{mygreen}{+1.1} & \scriptsize \bftable \textcolor{mygreen}{+0.7} & \scriptsize \bftable \textcolor{mygreen}{+1.9} & \scriptsize \bftable \textcolor{mygreen}{+1.1} & \scriptsize \bftable \textcolor{myred}{-1.1} & \scriptsize \bftable \textcolor{mygreen}{+0.8} \\
  \cmidrule(lr){2-18}
  & \colorcellgray \textit{Non-attention-based} & \colorcellgray & \colorcellgray & \colorcellgray & \colorcellgray & \colorcellgray & \colorcellgray & \colorcellgray & \colorcellgray & \colorcellgray & \colorcellgray & \colorcellgray & \colorcellgray & \colorcellgray & \colorcellgray & \colorcellgray & \colorcellgray \\
  & $\phi_{\text{KNN}}$ & 66\% & 23.9 & 15.1 & 24.9 & \underline{26.0} & \underline{29.6} & 58.4 & 39.9 & 60.9 & 74.4 & \bftable 58.4 & 62.8 & 81.2 & 55.9 & 61.5 & 52.8 \\
  & $\phi_{\text{uniform}}$ & 66\% & \bftable 28.0 & 20.6 & \bftable 28.6 & \bftable 29.7 & \bftable 33.3 & \underline{59.0} & 41.4 & \underline{61.8} & 75.9 & 57.1 & 64.6 & \underline{85.2} & 58.1 & 61.9 & 53.0 \\
   \cmidrule(lr){2-18}
  & \colorcellgray \textit{Ensemble} & \colorcellgray & \colorcellgray & \colorcellgray & \colorcellgray & \colorcellgray & \colorcellgray & \colorcellgray & \colorcellgray & \colorcellgray & \colorcellgray & \colorcellgray & \colorcellgray & \colorcellgray & \colorcellgray & \colorcellgray & \colorcellgray  \\
  & \colorcell $\phi_{\textbf{\textit{-R}}} + \phi_{\text{uniform}}$ (Ours) & \colorcell 61\% & \underline{\colorcell 27.2} & \bftable \colorcell 29.1 & \underline{\colorcell 27.2} & \colorcell 24.7 & \colorcell 27.7 & \bftable \colorcell 61.2 & \bftable \colorcell 46.6 & \bftable \colorcell 62.3 & \bftable \colorcell 77.4 & \bftable \colorcell 58.4 & \bftable \colorcell 65.4 & \bftable \colorcell 86.0 & \bftable \colorcell 58.9 & \underline{\colorcell 62.7} & \bftable \colorcell 54.0 \\
  \midrule
  \midrule
  & \colorcellgray \textit{Attention-based}& \colorcellgray & \colorcellgray & \colorcellgray & \colorcellgray & \colorcellgray & \colorcellgray & \colorcellgray & \colorcellgray & \colorcellgray & \colorcellgray & \colorcellgray & \colorcellgray & \colorcellgray & \colorcellgray & \colorcellgray & \colorcellgray \\
  \multirow{8}{*}{$K = 8$} & $\phi_{\text{original}}$ & 56\% & 23.3 & 19.4 & 23.5 & 24.0 & 26.3 & 59.8 & 45.0 & 60.3 & 76.1 & 57.8 & 64.6 & 85.4 & 57.5 & 62.6 & 53.0 \\
  & \colorcell $\phi_{\textbf{\textit{-R}}}$ (Ours) & \colorcell 56\% & \colorcell 27.3 & \underline{\colorcell 27.1} & \colorcell 26.7 & \colorcell 26.4 & \colorcell 29.2 & \underline{\colorcell 61.4} & \underline{\colorcell 49.0} & \colorcell 61.5 & \underline{\colorcell 77.4} & \colorcell 57.8 & \underline{\colorcell 65.5} & \underline{\colorcell 86.7} & \underline{\colorcell 58.6} & \underline{\colorcell 63.0} & \underline{\colorcell 53.7} \\
  & \scriptsize \textcolor{mygreen}{$\Delta$} & & \scriptsize \bftable \textcolor{mygreen}{+4.0} & \scriptsize \bftable \textcolor{mygreen}{+7.6} & \scriptsize \bftable \textcolor{mygreen}{+3.2} & \scriptsize \bftable \textcolor{mygreen}{+2.5} & \scriptsize \bftable \textcolor{mygreen}{+2.9} & \scriptsize \bftable \textcolor{mygreen}{+1.6} & \scriptsize \bftable \textcolor{mygreen}{+4.0} & \scriptsize \bftable \textcolor{mygreen}{+1.2} & \scriptsize \bftable \textcolor{mygreen}{+1.3} & \scriptsize \bftable \textcolor{gray}{+0.0} & \scriptsize \bftable \textcolor{mygreen}{+0.8} & \scriptsize \bftable \textcolor{mygreen}{+1.3} & \scriptsize \bftable \textcolor{mygreen}{+1.1} & \scriptsize \bftable \textcolor{mygreen}{+0.4} & \scriptsize \bftable \textcolor{mygreen}{+0.6} \\
  \cmidrule(lr){2-18}
  & \colorcellgray \textit{Non-attention-based} & \colorcellgray & \colorcellgray & \colorcellgray & \colorcellgray & \colorcellgray & \colorcellgray & \colorcellgray & \colorcellgray & \colorcellgray & \colorcellgray & \colorcellgray & \colorcellgray & \colorcellgray & \colorcellgray & \colorcellgray & \colorcellgray  \\
  & $\phi_{\text{KNN}}$ & 55\% & 23.6 & 15.4 & 24.4 & 25.2 & 29.4 & 58.6 & 40.2 & 61.1 & 74.5 & \underline{58.5} & 62.9 & 81.4 & 56.2 & 60.9 & 53.0 \\
  & $\phi_{\text{uniform}}$ & 55\% & \underline{30.3} & 24.6 & \underline{31.0} & \underline{30.9} & \underline{34.8} & 59.3 & 42.2 & \underline{61.8} & 76.0 & 57.4 & 64.4 & 85.3 & 57.9 & 61.0 & 53.2 \\
   \cmidrule(lr){2-18}
  & \colorcellgray \textit{Ensemble} & \colorcellgray & \colorcellgray & \colorcellgray & \colorcellgray & \colorcellgray & \colorcellgray & \colorcellgray & \colorcellgray & \colorcellgray & \colorcellgray & \colorcellgray & \colorcellgray & \colorcellgray & \colorcellgray & \colorcellgray & \colorcellgray \\
  & \colorcell$\phi_{\textbf{\textit{-R}}} + \phi_{\text{uniform}}$ (Ours) & \colorcell 50\% & \bftable \colorcell 35.6 & \bftable \colorcell 32.0 & \bftable \colorcell 35.9 & \bftable \colorcell 35.4 & \bftable \colorcell 38.8 & \bftable \colorcell 62.7 & \bftable \colorcell 51.7 & \bftable \colorcell 62.4 & \bftable \colorcell 78.1 & \bftable \colorcell 58.6 & \bftable \colorcell 66.0 & \bftable \colorcell 87.4 & \bftable \colorcell 59.1 & \bftable \colorcell 63.6 & \bftable \colorcell 54.0 \\

  \end{tabular}
\end{small}
\caption{Evaluating alternative criteria for token pruning after the early LLM layers. For each task and pruning layer, we \textbf{bold} the best result and \underline{underline} the second-best result. Our main findings include: (1) our proposed RoPE-free criteria $\phi_{\textbf{\textit{-R}}}$ substantially improves pruning performance compared to the original criteria $\phi_{\text{original}}$; (2) pruning later ($K=8$) yields higher performance than pruning earlier ($K=3$); and (3) integrating uniform sampling into the attention-based criteria with $\phi_{\textbf{\textit{-R}}} + \phi_{\text{uniform}}$ enhances effectiveness. Using $\textit{R} = 0.75$ for $\phi_{\text{original}}$, $\phi_{\textbf{\textit{-R}}}$, and $\phi_{\textbf{\textit{-R}}} + \phi_{\text{uniform}}$. See \S\ref{sec:improving_pruning_criteria} for criteria definitions.   
    }
  \label{tab:fastv_alternatives}
\end{table*}

\begin{table}[ht]
  \centering
 \begin{small}
  \begin{tabular}{lcccc}
    \toprule
    $K$ & OCID-Ref & RefCOCOg & RefCOCO+ & RefCOCO \\
    \midrule
    3 & 23.8 & 16.2 & 14.5 & 16.3 \\
    8 & 26.7 & 29.8 & 26.8 & 29.8 \\
    \bottomrule
  \end{tabular}
\end{small}
\caption{Assessing localization performance when using the selected tokens from $\phi_{\textbf{\textit{-R}}}$ for the entirety of the LLM. We find that $\phi_{\textbf{\textit{-R}}}$ applied at a later LLM layer results in a better token selection.}
\label{tab:rope_free_oracle}
\end{table}

\textbf{Token selection improves when pruning later:} Comparing $\phi_{\textbf{\textit{-R}}}$ when $K=8$ versus $K=3$, we see that the average localization performance improves by 63\%. However, it remains unclear whether this performance increase is affected by factors other than the ability of the criteria to select important tokens. Therefore, to directly compare the criteria across different layers, we take the tokens selected by both criteria and evaluate how well the model can perform with only these tokens passed into the LLM (as is done in \S\ref{sec:vlm_inference_on_other_tasks}). As shown in Table~\ref{tab:rope_free_oracle}, we find the selected tokens from $\phi_{\textbf{\textit{-R}}}$ applied after layer 8 are superior compared the tokens from $\phi_{\textbf{\textit{-R}}}$ applied after layer 3.

\textbf{Integrating uniform sampling with attention-based criteria is beneficial in early layers:} When comparing the attention-based criteria $\phi_{\textbf{\textit{-R}}}$ with the two non-attention based criteria $\phi_{\text{KNN}}$ and $\phi_{\text{uniform}}$, we find that $\phi_{\textbf{\textit{-R}}}$ results in better performance on some tasks (e.g., OCID-Ref, TextVQA), while $\phi_{\text{KNN}}$ and $\phi_{\text{uniform}}$ outperform on other tasks (e.g., RefCOCO). This varying effectiveness of criteria types presumably comes from some tasks requiring a more detailed understanding of specific image regions while others benefit from a broader understanding with full image coverage.
When we combine these two criteria types with $\phi_{\textbf{\textit{-R}}} + \phi_{\text{uniform}}$, we find that when $K=3$ this approach improves upon localization performance of $\phi_{\textbf{\textit{-R}}}$ by 63\% but has a slight decrease compared to $\phi_{\text{uniform}}$ by 2.9\%. However, when $K=8$, this approach far outperforms both of the individual criteria, improving $\phi_{\textbf{\textit{-R}}}$ by 30\% and $\phi_{\text{uniform}}$ by 17\%. Note that this setup results in slightly less computation efficiency than $\phi_{\textbf{\textit{-R}}}$ (7\% drop in FLOPS reduction for $K=3$ and 6\% for $K=16$), but this is outweighed by the performance improvements.

We provide qualitative examples of various criteria token selection and additional results using a different model setup in the supplement.

\begin{figure*}[th]
  \centering
   \includegraphics[width=1\linewidth]{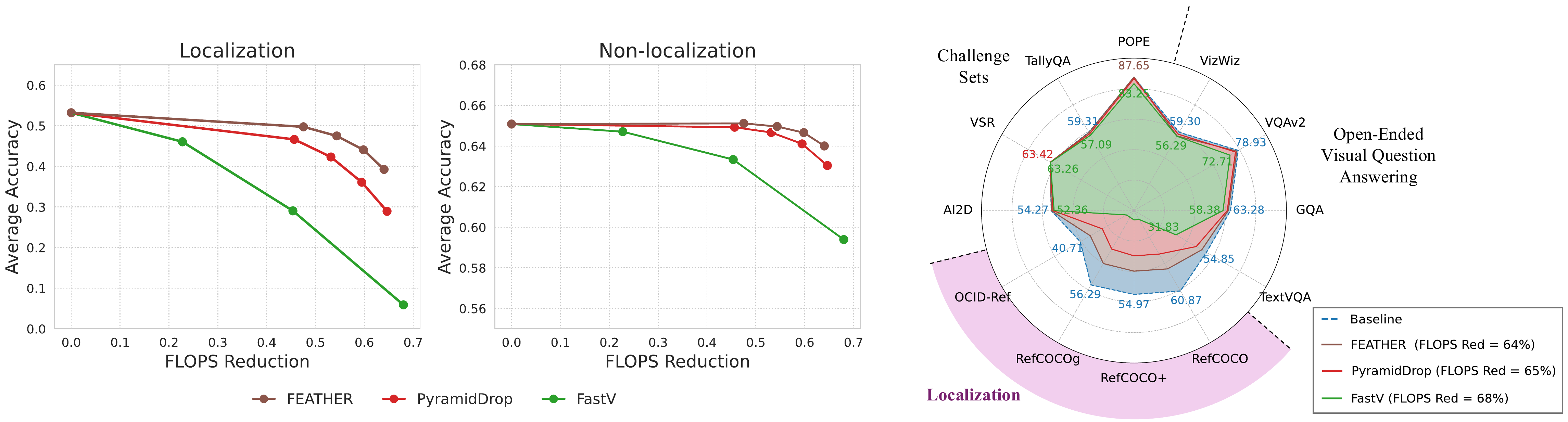}
   \caption{Comparing \feather performance against FastV \cite{chen2024fastv} and PyramidDrop \cite{xing2024pyramiddrop}. We find that \feather far outperforms both compared methods, particularly for the vision-centric task of localization.}
   \label{fig:feather_results}
\end{figure*}

\subsection{Distilling insights}
\label{sec:distilling_insights}

Guided by our insights, we now present our final approach \feather (\featherlong). In this approach, we first perform pruning after an early layer ($K = 8$), utilizing our proposed criteria $\phi_{\textbf{\textit{-R}}} + \phi_{\text{uniform}}$ to retain $(1-R)\%$ of tokens. Given our finding that the criteria improves as the LLM layer increases, we additionally prune a second time at $K = 16$. At this stage, we utilize $\phi_{\textbf{\textit{-R}}}$ as the criteria, as uniform sampling should not be necessary when the attention-based criteria reliably selects important tokens.
For the reduction ratio, we choose to only retain $(1-R)^2\%$ of the remaining tokens since the attention-based criteria has proved highly effective when pruning at later layers, even when using the original criteria $\phi_{\text{original}}$ (see Figure~\ref{fig:fastv_ineffective_criteria}(c)).

\subsection{Comparison against FastV and PyramidDrop}
\label{sec:feather_results}

We compare our approach against the one-stage early pruning approach of FastV ($K = 3$) \cite{chen2024fastv} and the multi-stage approach of PyramidDrop ($K = [8, 16, 24]$) \cite{xing2024pyramiddrop}. Note that both of these methods use $\phi_{\text{original}}$ for the criteria. 

As shown in Figure~\ref{fig:feather_results}, we find that \feather far outperforms the baselines, especially on the localization tasks. Specifically, for comparable computational costs (64\% FLOPS reduction for our approach, 68\% FLOPS reduction for FastV, and 65\% FLOPS reduction for PyramidDrop), we observe that for localization tasks,  \feather exhibits more than $\mathbf{5\times}$ average performance improvement compared to FastV and a 36\% average performance improvement compared to PyramidDrop. For non-localization tasks, \feather has a 7.8\% improvement over FastV and 1.5\% improvement over PyramidDrop. Note that PyramidDrop performs substantially better than FastV as it prunes fewer tokens in an early layer. However, it still suffers from an ineffective pruning strategy at this stage, though the impact is less pronounced since it predominantly prunes later.

Remarkably, with our 64\% FLOPS reduction setup, after layer 16 only 3.3\% of tokens are retained, yet the average localization performance decrease compared to the baseline method with no token pruning is only 26\% (Figure~\ref{fig:pull_fig} includes an example of retained tokens after layer 16). This finding illustrates that even for vision-centric tasks, maintaining strong performance while gaining huge acceleration speedups with extensive pruning is possible, but it heavily relies on the effectiveness of the pruning criteria.

\section{Conclusion}
\label{sec:discussion}

In this work, we examine the visual capabilities of the VLM acceleration approach of pruning visual tokens after shallow LLM layers. While strong performance is maintained on most evaluated tasks, it fails on more vision-centric tasks like TextVQA and localization due to its flawed pruning criteria that predominately selects visual tokens from the bottom part of the image. Observing this same flawed criteria on other tasks, we show that strong performance is largely due to the benchmarks' inability to assess fine-grained visual capabilities. Next, we propose and evaluate several alternative criteria to improve visual capabilities, ultimately arriving at our final method, \feather.
This approach refines the attention-based criteria to address the token selection bias while using uniform sampling for better image coverage. It then prunes more aggressively when the criteria more effectively identifies important tokens.
We find that \feather has more than $\mathbf{5\times}$ performance improvement on localization compared to the original acceleration approach.

While we study a particular type of VLM acceleration, our work highlights a broad challenge in evaluating VLMs. Building on \cite{tong2024cambrian}, which shows that text-only models can perform well on some vision-language benchmarks, we demonstrate that even benchmarks with substantial differences between vision-enabled and disabled setups may not assess fine-grained visual capabilities. We address this issue by focusing on more vision-centric localization tasks, though this limits the analysis to a few specific types of skills.
To accurately assess a wider range of visual capabilities, future work may explore how to resolve current dataset biases that models can exploit \cite{li2025naturalbench}. 
Additionally, while removing RoPE for token pruning effectively eliminates positional bias, it may introduce unintended effects on attention weights influencing token selection. Future research could investigate more robust methods for encoding positional information in visually conditioned language models to prevent positional artifacts in cross-modal interactions.

\clearpage
\noindent \textbf{Acknowledgments.} This work is supported in part by the National Science Foundation (NSF) under Grant No. 2026498, the Stanford AIMI-HAI Partnership Grant, and the NSF Graduate Research Fellowship Program under Grant No. DGE-2146755 (for M.E.). Any opinions, findings, and conclusions or recommendations expressed in this material are those of the authors and do not necessarily reflect the views of any other entity.
{
    \small
    \bibliographystyle{ieeenat_fullname}
    \bibliography{main}

\begin{thebibliography}{44}
\providecommand{\natexlab}[1]{#1}
\providecommand{\url}[1]{\texttt{#1}}
\expandafter\ifx\csname urlstyle\endcsname\relax
  \providecommand{\doi}[1]{doi: #1}\else
  \providecommand{\doi}{doi: \begingroup \urlstyle{rm}\Url}\fi

\bibitem[Acharya et~al.(2019)Acharya, Kafle, and Kanan]{acharya2019tallyqa}
Manoj Acharya, Kushal Kafle, and Christopher Kanan.
\newblock Tallyqa: Answering complex counting questions.
\newblock In \emph{Proceedings of the AAAI conference on artificial intelligence}, pages 8076--8084, 2019.

\bibitem[Alayrac et~al.(2022)Alayrac, Donahue, Luc, Miech, Barr, Hasson, Lenc, Mensch, Millican, Reynolds, et~al.]{alayrac2022flamingo}
Jean-Baptiste Alayrac, Jeff Donahue, Pauline Luc, Antoine Miech, Iain Barr, Yana Hasson, Karel Lenc, Arthur Mensch, Katherine Millican, Malcolm Reynolds, et~al.
\newblock Flamingo: a visual language model for few-shot learning.
\newblock \emph{Advances in neural information processing systems}, 35:\penalty0 23716--23736, 2022.

\bibitem[Arif et~al.(2024)Arif, Yoon, Nikolopoulos, Vandierendonck, John, and Ji]{arif2024hired}
Kazi Hasan~Ibn Arif, JinYi Yoon, Dimitrios~S Nikolopoulos, Hans Vandierendonck, Deepu John, and Bo Ji.
\newblock Hired: Attention-guided token dropping for efficient inference of high-resolution vision-language models in resource-constrained environments.
\newblock \emph{arXiv preprint arXiv:2408.10945}, 2024.

\bibitem[Bai et~al.(2023)Bai, Bai, Yang, Wang, Tan, Wang, Lin, Zhou, and Zhou]{bai2023qwenvl}
Jinze Bai, Shuai Bai, Shusheng Yang, Shijie Wang, Sinan Tan, Peng Wang, Junyang Lin, Chang Zhou, and Jingren Zhou.
\newblock Qwen-vl: A frontier large vision-language model with versatile abilities.
\newblock \emph{arXiv preprint arXiv:2308.12966}, 2023.

\bibitem[Bigham et~al.(2010)Bigham, Jayant, Ji, Little, Miller, Miller, Miller, Tatarowicz, White, White, et~al.]{bigham2010vizwiz}
Jeffrey~P Bigham, Chandrika Jayant, Hanjie Ji, Greg Little, Andrew Miller, Robert~C Miller, Robin Miller, Aubrey Tatarowicz, Brandyn White, Samual White, et~al.
\newblock Vizwiz: nearly real-time answers to visual questions.
\newblock In \emph{Proceedings of the 23nd annual ACM symposium on User interface software and technology}, pages 333--342, 2010.

\bibitem[Bolya et~al.(2023)Bolya, Fu, Dai, Zhang, Feichtenhofer, and Hoffman]{bolya2023tokenmerging}
Daniel Bolya, Cheng-Yang Fu, Xiaoliang Dai, Peizhao Zhang, Christoph Feichtenhofer, and Judy Hoffman.
\newblock Token merging: Your vit but faster.
\newblock In \emph{The Eleventh International Conference on Learning Representations}, 2023.

\bibitem[Cai et~al.(2024)Cai, Yang, Gao, and Lee]{cai2024matryoshka}
Mu Cai, Jianwei Yang, Jianfeng Gao, and Yong~Jae Lee.
\newblock Matryoshka multimodal models.
\newblock \emph{arXiv preprint arXiv:2405.17430}, 2024.

\bibitem[Chai et~al.(2024)Chai, Song, Du, Meng, Madhavan, Bar-Tal, Hwang, Xie, and Manning]{chai2024auroracap}
Wenhao Chai, Enxin Song, Yilun Du, Chenlin Meng, Vashisht Madhavan, Omer Bar-Tal, Jeng-Neng Hwang, Saining Xie, and Christopher~D Manning.
\newblock Auroracap: Efficient, performant video detailed captioning and a new benchmark.
\newblock \emph{arXiv preprint arXiv:2410.03051}, 2024.

\bibitem[Chen et~al.(2024{\natexlab{a}})Chen, Ye, He, Wang, Khashabi, and Yuille]{chen2024llavolta}
Jieneng Chen, Luoxin Ye, Ju He, Zhao-Yang Wang, Daniel Khashabi, and Alan Yuille.
\newblock Efficient large multi-modal models via visual context compression.
\newblock In \emph{The Thirty-eighth Annual Conference on Neural Information Processing Systems}, 2024{\natexlab{a}}.

\bibitem[Chen et~al.(2024{\natexlab{b}})Chen, Zhao, Liu, Bai, Lin, Zhou, and Chang]{chen2024fastv}
Liang Chen, Haozhe Zhao, Tianyu Liu, Shuai Bai, Junyang Lin, Chang Zhou, and Baobao Chang.
\newblock An image is worth 1/2 tokens after layer 2: Plug-and-play inference acceleration for large vision-language models.
\newblock \emph{ECCV}, 2024{\natexlab{b}}.

\bibitem[Chen et~al.(2024{\natexlab{c}})Chen, Xu, Zhang, Liu, Liu, and Liu]{chen2024recoverablecompression}
Yi Chen, Jian Xu, Xu-Yao Zhang, Wen-Zhuo Liu, Yang-Yang Liu, and Cheng-Lin Liu.
\newblock Recoverable compression: A multimodal vision token recovery mechanism guided by text information.
\newblock \emph{arXiv preprint arXiv:2409.01179}, 2024{\natexlab{c}}.

\bibitem[Dao et~al.(2022)Dao, Fu, Ermon, Rudra, and R{\'e}]{dao2022flashattention}
Tri Dao, Dan Fu, Stefano Ermon, Atri Rudra, and Christopher R{\'e}.
\newblock Flashattention: Fast and memory-efficient exact attention with io-awareness.
\newblock \emph{Advances in neural information processing systems}, 35:\penalty0 16344--16359, 2022.

\bibitem[Goyal et~al.(2017)Goyal, Khot, Summers-Stay, Batra, and Parikh]{goyal2017vqav2}
Yash Goyal, Tejas Khot, Douglas Summers-Stay, Dhruv Batra, and Devi Parikh.
\newblock Making the v in vqa matter: Elevating the role of image understanding in visual question answering.
\newblock In \emph{Proceedings of the IEEE conference on computer vision and pattern recognition}, pages 6904--6913, 2017.

\bibitem[He et~al.(2024)He, Chen, Liu, Shao, Zhou, Zhang, and Zhuang]{he2024zipvl}
Yefei He, Feng Chen, Jing Liu, Wenqi Shao, Hong Zhou, Kaipeng Zhang, and Bohan Zhuang.
\newblock Zipvl: Efficient large vision-language models with dynamic token sparsification and kv cache compression.
\newblock \emph{arXiv preprint arXiv:2410.08584}, 2024.

\bibitem[Hong et~al.(2024)Hong, Jiang, Qi, Meng, Yu, Zhou, and Zhou]{hong2024token}
Xiangyu Hong, Che Jiang, Biqing Qi, Fandong Meng, Mo Yu, Bowen Zhou, and Jie Zhou.
\newblock On the token distance modeling ability of higher rope attention dimension.
\newblock \emph{arXiv preprint arXiv:2410.08703}, 2024.

\bibitem[Hudson and Manning(2019)]{hudson2019gqa}
Drew~A Hudson and Christopher~D Manning.
\newblock Gqa: A new dataset for real-world visual reasoning and compositional question answering.
\newblock In \emph{Proceedings of the IEEE/CVF conference on computer vision and pattern recognition}, pages 6700--6709, 2019.

\bibitem[Jin et~al.(2024)Jin, Takanobu, Zhang, Cao, and Yuan]{jin2024chatunivi}
Peng Jin, Ryuichi Takanobu, Wancai Zhang, Xiaochun Cao, and Li Yuan.
\newblock Chat-univi: Unified visual representation empowers large language models with image and video understanding.
\newblock In \emph{Proceedings of the IEEE/CVF Conference on Computer Vision and Pattern Recognition}, pages 13700--13710, 2024.

\bibitem[Karamcheti et~al.(2024)Karamcheti, Nair, Balakrishna, Liang, Kollar, and Sadigh]{karamcheti2024prismatic}
Siddharth Karamcheti, Suraj Nair, Ashwin Balakrishna, Percy Liang, Thomas Kollar, and Dorsa Sadigh.
\newblock Prismatic {VLM}s: Investigating the design space of visually-conditioned language models.
\newblock In \emph{Proceedings of the 41st International Conference on Machine Learning}, pages 23123--23144. PMLR, 2024.

\bibitem[Kazemzadeh et~al.(2014)Kazemzadeh, Ordonez, Matten, and Berg]{kazemzadeh2014referitgame}
Sahar Kazemzadeh, Vicente Ordonez, Mark Matten, and Tamara Berg.
\newblock Referitgame: Referring to objects in photographs of natural scenes.
\newblock In \emph{Proceedings of the 2014 conference on empirical methods in natural language processing (EMNLP)}, pages 787--798, 2014.

\bibitem[Kembhavi et~al.(2016)Kembhavi, Salvato, Kolve, Seo, Hajishirzi, and Farhadi]{kembhavi2016ai2d}
Aniruddha Kembhavi, Mike Salvato, Eric Kolve, Minjoon Seo, Hannaneh Hajishirzi, and Ali Farhadi.
\newblock A diagram is worth a dozen images.
\newblock In \emph{Computer Vision--ECCV 2016: 14th European Conference, Amsterdam, The Netherlands, October 11--14, 2016, Proceedings, Part IV 14}, pages 235--251. Springer, 2016.

\bibitem[Li et~al.(2025)Li, Lin, Peng, Nyandwi, Jiang, Ma, Khanuja, Krishna, Neubig, and Ramanan]{li2025naturalbench}
Baiqi Li, Zhiqiu Lin, Wenxuan Peng, Jean de~Dieu Nyandwi, Daniel Jiang, Zixian Ma, Simran Khanuja, Ranjay Krishna, Graham Neubig, and Deva Ramanan.
\newblock Naturalbench: Evaluating vision-language models on natural adversarial samples.
\newblock \emph{Advances in Neural Information Processing Systems}, 37:\penalty0 17044--17068, 2025.

\bibitem[Li et~al.(2022)Li, Li, Xiong, and Hoi]{li2022blip}
Junnan Li, Dongxu Li, Caiming Xiong, and Steven Hoi.
\newblock Blip: Bootstrapping language-image pre-training for unified vision-language understanding and generation.
\newblock In \emph{International conference on machine learning}, pages 12888--12900. PMLR, 2022.

\bibitem[Li et~al.(2023{\natexlab{a}})Li, Li, Savarese, and Hoi]{li2023blip2}
Junnan Li, Dongxu Li, Silvio Savarese, and Steven Hoi.
\newblock Blip-2: Bootstrapping language-image pre-training with frozen image encoders and large language models.
\newblock In \emph{International conference on machine learning}, pages 19730--19742. PMLR, 2023{\natexlab{a}}.

\bibitem[Li et~al.(2024)Li, Goyal, Semedo, and Kolter]{li2024onetoken}
Kevin~Y Li, Sachin Goyal, Joao~D Semedo, and J~Zico Kolter.
\newblock Inference optimal vlms need only one visual token but larger models.
\newblock \emph{arXiv preprint arXiv:2411.03312}, 2024.

\bibitem[Li et~al.(2023{\natexlab{b}})Li, Du, Zhou, Wang, Zhao, and Wen]{li2023pope}
Yifan Li, Yifan Du, Kun Zhou, Jinpeng Wang, Wayne~Xin Zhao, and Ji-Rong Wen.
\newblock Evaluating object hallucination in large vision-language models.
\newblock \emph{arXiv preprint arXiv:2305.10355}, 2023{\natexlab{b}}.

\bibitem[Liu et~al.(2023)Liu, Emerson, and Collier]{liu2023vsr}
Fangyu Liu, Guy Emerson, and Nigel Collier.
\newblock Visual spatial reasoning.
\newblock \emph{Transactions of the Association for Computational Linguistics}, 11:\penalty0 635--651, 2023.

\bibitem[Liu et~al.(2024{\natexlab{a}})Liu, Li, Li, and Lee]{liu2024llava1.5}
Haotian Liu, Chunyuan Li, Yuheng Li, and Yong~Jae Lee.
\newblock Improved baselines with visual instruction tuning.
\newblock In \emph{Proceedings of the IEEE/CVF Conference on Computer Vision and Pattern Recognition}, pages 26296--26306, 2024{\natexlab{a}}.

\bibitem[Liu et~al.(2024{\natexlab{b}})Liu, Li, Li, Li, Zhang, Shen, and Lee]{liu2024llavanext}
Haotian Liu, Chunyuan Li, Yuheng Li, Bo Li, Yuanhan Zhang, Sheng Shen, and Yong~Jae Lee.
\newblock Llava-next: Improved reasoning, ocr, and world knowledge, 2024{\natexlab{b}}.

\bibitem[Liu et~al.(2024{\natexlab{c}})Liu, Li, Wu, and Lee]{liu2024llava}
Haotian Liu, Chunyuan Li, Qingyang Wu, and Yong~Jae Lee.
\newblock Visual instruction tuning.
\newblock \emph{Advances in neural information processing systems}, 36, 2024{\natexlab{c}}.

\bibitem[Liu et~al.(2024{\natexlab{d}})Liu, Gehrig, Messikommer, Cannici, and Scaramuzza]{liu2024revisitingtokenpruning}
Yifei Liu, Mathias Gehrig, Nico Messikommer, Marco Cannici, and Davide Scaramuzza.
\newblock Revisiting token pruning for object detection and instance segmentation.
\newblock In \emph{Proceedings of the IEEE/CVF Winter Conference on Applications of Computer Vision}, pages 2658--2668, 2024{\natexlab{d}}.

\bibitem[Shang et~al.(2024)Shang, Cai, Xu, Lee, and Yan]{shang2024prumerge}
Yuzhang Shang, Mu Cai, Bingxin Xu, Yong~Jae Lee, and Yan Yan.
\newblock Llava-prumerge: Adaptive token reduction for efficient large multimodal models.
\newblock \emph{arXiv preprint arXiv:2403.15388}, 2024.

\bibitem[Singh et~al.(2019)Singh, Natarajan, Shah, Jiang, Chen, Batra, Parikh, and Rohrbach]{singh2019textvqa}
Amanpreet Singh, Vivek Natarajan, Meet Shah, Yu Jiang, Xinlei Chen, Dhruv Batra, Devi Parikh, and Marcus Rohrbach.
\newblock Towards vqa models that can read.
\newblock In \emph{Proceedings of the IEEE/CVF conference on computer vision and pattern recognition}, pages 8317--8326, 2019.

\bibitem[Su et~al.(2024)Su, Ahmed, Lu, Pan, Bo, and Liu]{su2024roformer}
Jianlin Su, Murtadha Ahmed, Yu Lu, Shengfeng Pan, Wen Bo, and Yunfeng Liu.
\newblock Roformer: Enhanced transformer with rotary position embedding.
\newblock \emph{Neurocomputing}, 568:\penalty0 127063, 2024.

\bibitem[Tong et~al.(2024)Tong, Brown, Wu, Woo, Middepogu, Akula, Yang, Yang, Iyer, Pan, et~al.]{tong2024cambrian}
Shengbang Tong, Ellis Brown, Penghao Wu, Sanghyun Woo, Manoj Middepogu, Sai~Charitha Akula, Jihan Yang, Shusheng Yang, Adithya Iyer, Xichen Pan, et~al.
\newblock Cambrian-1: A fully open, vision-centric exploration of multimodal llms.
\newblock \emph{arXiv preprint arXiv:2406.16860}, 2024.

\bibitem[Touvron et~al.(2023)Touvron, Martin, Stone, Albert, Almahairi, Babaei, Bashlykov, Batra, Bhargava, Bhosale, et~al.]{touvron2023llama2}
Hugo Touvron, Louis Martin, Kevin Stone, Peter Albert, Amjad Almahairi, Yasmine Babaei, Nikolay Bashlykov, Soumya Batra, Prajjwal Bhargava, Shruti Bhosale, et~al.
\newblock Llama 2: Open foundation and fine-tuned chat models.
\newblock \emph{arXiv preprint arXiv:2307.09288}, 2023.

\bibitem[Wan et~al.(2024)Wan, Wu, Liu, Huang, Zhu, Jin, Wang, and Yuan]{wan2024look}
Zhongwei Wan, Ziang Wu, Che Liu, Jinfa Huang, Zhihong Zhu, Peng Jin, Longyue Wang, and Li Yuan.
\newblock Look-m: Look-once optimization in kv cache for efficient multimodal long-context inference.
\newblock \emph{arXiv preprint arXiv:2406.18139}, 2024.

\bibitem[Wang et~al.(2021)Wang, Liu, Su, Wang, Wang, Hsu, and Chen]{wang2021-ocidref}
Ke-Jyun Wang, Yun-Hsuan Liu, Hung-Ting Su, Jen-Wei Wang, Yu-Siang Wang, Winston Hsu, and Wen-Chin Chen.
\newblock {OCID}-ref: A 3{D} robotic dataset with embodied language for clutter scene grounding.
\newblock In \emph{Proceedings of the 2021 Conference of the North American Chapter of the Association for Computational Linguistics: Human Language Technologies}, pages 5333--5338, Online, 2021. Association for Computational Linguistics.

\bibitem[Wang et~al.(2024)Wang, Bai, Tan, Wang, Fan, Bai, Chen, Liu, Wang, Ge, et~al.]{wang2024qwen2vl}
Peng Wang, Shuai Bai, Sinan Tan, Shijie Wang, Zhihao Fan, Jinze Bai, Keqin Chen, Xuejing Liu, Jialin Wang, Wenbin Ge, et~al.
\newblock Qwen2-vl: Enhancing vision-language model's perception of the world at any resolution.
\newblock \emph{arXiv preprint arXiv:2409.12191}, 2024.

\bibitem[Xing et~al.(2024)Xing, Huang, Dong, Lu, Zhang, Zang, Cao, He, Wang, Wu, et~al.]{xing2024pyramiddrop}
Long Xing, Qidong Huang, Xiaoyi Dong, Jiajie Lu, Pan Zhang, Yuhang Zang, Yuhang Cao, Conghui He, Jiaqi Wang, Feng Wu, et~al.
\newblock Pyramiddrop: Accelerating your large vision-language models via pyramid visual redundancy reduction.
\newblock \emph{arXiv preprint arXiv:2410.17247}, 2024.

\bibitem[Yang et~al.(2025)Yang, Chen, Tian, Wang, Li, Yu, and Jia]{yang2025visionzip}
Senqiao Yang, Yukang Chen, Zhuotao Tian, Chengyao Wang, Jingyao Li, Bei Yu, and Jiaya Jia.
\newblock Visionzip: Longer is better but not necessary in vision language models.
\newblock In \emph{Proceedings of the Computer Vision and Pattern Recognition Conference}, pages 19792--19802, 2025.

\bibitem[Yu et~al.(2024)Yu, Chen, and Xu]{yu2024balancingperformanceefficiency}
Gaotong Yu, Yi Chen, and Jian Xu.
\newblock Balancing performance and efficiency: A multimodal large language model pruning method based image text interaction.
\newblock \emph{arXiv preprint arXiv:2409.01162}, 2024.

\bibitem[Yu et~al.(2016)Yu, Poirson, Yang, Berg, and Berg]{yu2016refcoco}
Licheng Yu, Patrick Poirson, Shan Yang, Alexander~C Berg, and Tamara~L Berg.
\newblock Modeling context in referring expressions.
\newblock In \emph{Computer Vision--ECCV 2016: 14th European Conference, Amsterdam, The Netherlands, October 11-14, 2016, Proceedings, Part II 14}, pages 69--85. Springer, 2016.

\bibitem[Zhai et~al.(2023)Zhai, Mustafa, Kolesnikov, and Beyer]{zhai2023siglip}
Xiaohua Zhai, Basil Mustafa, Alexander Kolesnikov, and Lucas Beyer.
\newblock Sigmoid loss for language image pre-training.
\newblock In \emph{Proceedings of the IEEE/CVF International Conference on Computer Vision}, pages 11975--11986, 2023.

\bibitem[Zhang et~al.(2024)Zhang, Cheng, Lu, Zhuo, Wang, Cao, Guo, She, and Zhang]{zhang2024fastervlm}
Qizhe Zhang, Aosong Cheng, Ming Lu, Zhiyong Zhuo, Minqi Wang, Jiajun Cao, Shaobo Guo, Qi She, and Shanghang Zhang.
\newblock [cls] attention is all you need for training-free visual token pruning: Make vlm inference faster.
\newblock \emph{arXiv e-prints}, pages arXiv--2412, 2024.

\end{thebibliography}
}

\clearpage

\maketitlesupplementary

\renewcommand{\thesection}{A\arabic{section}}
\renewcommand{\thefigure}{A\arabic{figure}}
\renewcommand{\thetable}{A\arabic{table}}

\setcounter{section}{0}
\setcounter{figure}{0}
\setcounter{table}{0}

\section{Results on Different Model Setup}
We additionally experiment with using a DINOv2 + SigLIP visual encoder. As shown in Table~\ref{tab:dinosiglip_results}, we observe the same behavior that removing RoPE substantially improves performance and incorporating uniform sampling is strong.

\section{Additional \feather Results}

We compare \feather performance against FastV and PyramidDrop on all evaluated benchmarks in Table~\ref{tab:feather_table}.

\vspace{2em}
\begin{table}[h]
  \centering
  \begin{minipage}{\textwidth}
  \centering
 \begin{small}
  \begin{tabular}{r@{\colpad}c@{\colpad}|@{\colpad}c@{\colpad}c@{\colpad}c@{\colpad}c@{\colpad}c@{\colpad}|@{\colpad}c@{\colpad}c@{\colpad}c@{\colpad}c@{\colpad}c@{\colpad}|@{\colpad}c@{\colpad}c@{\colpad}c@{\colpad}c@{\colpad}c@{\colpad}}
  \multicolumn{2}{c}{} & \multicolumn{5}{c}{Localization} & \multicolumn{5}{c}{Open-Ended VQA} & \multicolumn{5}{c}{Challenge Sets} \\
   Criteria & \rotatecol{FLOPS Red \vphantom{X}} & \rotatecol{Avg} & \rotatecol{OCID-Ref} & \rotatecol{RefCOCOg} & \rotatecol{RefCOCO+} & \rotatecol{RefCOCO} & \rotatecol{Avg} & \rotatecol{TextVQA} & \rotatecol{GQA} & \rotatecol{VQAv2} & \rotatecol{VizWiz} & \rotatecol{Avg}  & \rotatecol{POPE} & \rotatecol{TallyQA} & \rotatecol{VSR} & \rotatecol{AI2D} \\
  \midrule
   \colorcellgray \textit{Attention-based} & \colorcellgray & \colorcellgray & \colorcellgray & \colorcellgray & \colorcellgray & \colorcellgray & \colorcellgray & \colorcellgray & \colorcellgray & \colorcellgray & \colorcellgray & \colorcellgray & \colorcellgray & \colorcellgray & \colorcellgray & \colorcellgray \\
  $\phi_{\text{original}}$ & 68\% & 27.2 & 21.9 & 27.7 & 27.8 & 31.1 & 56.6 & 35.6 & 59.1 & 74.0 & 57.7 & 66.1 & 84.6 & 60.2 & \bftable 67.1 & 52.7 \\
  $\phi_{\textbf{\textit{-R}}}$ & 68\% & 37.2 & \underline{37.0} & 38.7 & 34.9 & 38.1 & \underline{60.1} & \underline{45.4} & 60.4 & \underline{76.5} & \underline{58.2} & \underline{66.3} & \underline{85.9} & \underline{61.1} & 65.1 & \underline{52.9} \\
  \scriptsize \textcolor{mygreen}{$\Delta$} & & \scriptsize \bftable \textcolor{mygreen}{+10.0} & \scriptsize \bftable \textcolor{mygreen}{+15.0} & \scriptsize \bftable \textcolor{mygreen}{+11.0} & \scriptsize \bftable \textcolor{mygreen}{+7.0} & \scriptsize \bftable \textcolor{mygreen}{+7.0} & \scriptsize \bftable \textcolor{mygreen}{+3.5} & \scriptsize \bftable \textcolor{mygreen}{+9.7} & \scriptsize \bftable \textcolor{mygreen}{+1.2} & \scriptsize \bftable \textcolor{mygreen}{+2.5} & \scriptsize \bftable \textcolor{mygreen}{+0.5} & \scriptsize \bftable \textcolor{mygreen}{+0.1} & \scriptsize \bftable \textcolor{mygreen}{+1.3} & \scriptsize \bftable \textcolor{mygreen}{+0.9} & \scriptsize \bftable \textcolor{myred}{-2.0} & \scriptsize \bftable \textcolor{mygreen}{+0.2}  \\
  \cmidrule(lr){1-17}
  \colorcellgray \textit{Non-attention-based} & \colorcellgray & \colorcellgray & \colorcellgray & \colorcellgray & \colorcellgray & \colorcellgray & \colorcellgray & \colorcellgray & \colorcellgray & \colorcellgray & \colorcellgray & \colorcellgray & \colorcellgray & \colorcellgray & \colorcellgray & \colorcellgray \\
  $\phi_{\text{KNN}}$ & 66\% & 20.5 & 13.4 & 22.1 & 22.0 & 24.6 & 54.2 & 29.9 & 60.0 & 70.2 & 57.0 & 60.5 & 77.7 & 51.9 & 61.9 & 50.7 \\
  $\phi_{\text{uniform}}$ & 66\% & \underline{38.3} & 32.7 & \underline{38.8} & \underline{38.8} & \underline{42.7} & 58.3 & 37.6 & \underline{61.9} & 75.8 & 58.0 & 65.8 & \underline{85.9} & 60.2 & 65.0 & 52.2 \\
   \cmidrule(lr){1-17}
  \colorcellgray \textit{Ensemble} & \colorcellgray & \colorcellgray & \colorcellgray & \colorcellgray & \colorcellgray & \colorcellgray & \colorcellgray & \colorcellgray & \colorcellgray & \colorcellgray & \colorcellgray & \colorcellgray & \colorcellgray & \colorcellgray & \colorcellgray & \colorcellgray  \\
  \colorcell $\phi_{\textbf{\textit{-R}}} + \phi_{\text{uniform}}$ (Ours) & \bftable \colorcell 61\% & \bftable \colorcell 46.3 & \bftable \colorcell 41.6 & \bftable \colorcell 47.3 & \bftable \colorcell 46.0 & \bftable \colorcell 50.1 & \bftable \colorcell 61.3 & \bftable \colorcell 46.8 & \bftable \colorcell 62.0 & \bftable \colorcell 77.7 & \bftable \colorcell 58.7 & \bftable \colorcell 66.8 & \bftable \colorcell 86.9 & \bftable \colorcell 61.6 & \colorcell \underline{65.4} & \bftable \colorcell 53.3 \\
  
  \end{tabular}
\end{small}
\caption{Evaluating criteria using DINOv2 + SigLIP visual encoder. For each task, we \textbf{bold} the best result and \underline{underline} the second-best result. Using $K=3$ for all setups.  
    }
  \label{tab:dinosiglip_results}
  \end{minipage}
\end{table}

\begin{table}[h]
  \centering
  \begin{minipage}{\textwidth}
  \centering
 \begin{small}
  \begin{tabular}{r@{\colpad}c@{\colpad}c@{\colpad}|@{\colpad}c@{\colpad}c@{\colpad}c@{\colpad}c@{\colpad}c@{\colpad}|@{\colpad}c@{\colpad}c@{\colpad}c@{\colpad}c@{\colpad}c@{\colpad}|@{\colpad}c@{\colpad}c@{\colpad}c@{\colpad}c@{\colpad}c@{\colpad}}
  \multicolumn{3}{c}{} & \multicolumn{5}{c}{Localization} & \multicolumn{5}{c}{Open-Ended VQA} & \multicolumn{5}{c}{Challenge Sets} \\
  Method & \rotatecol{FLOPS Red \vphantom{X}} & \rotatecol{GPU Hours} & \rotatecol{Avg} & \rotatecol{OCID-Ref} & \rotatecol{RefCOCOg} & \rotatecol{RefCOCO+} & \rotatecol{RefCOCO} & \rotatecol{Avg} & \rotatecol{TextVQA} & \rotatecol{GQA} & \rotatecol{VQAv2} & \rotatecol{VizWiz} & \rotatecol{Avg}  & \rotatecol{POPE} & \rotatecol{TallyQA} & \rotatecol{VSR} & \rotatecol{AI2D} \\
  \midrule
  Baseline & 0\% & 20.3 & 53.2 & 40.7 & 56.3 & 55.0 & 60.9 & 64.1 & 54.9 & 63.3 & 78.9 & 59.3 & 66.1 & 87.4 & 59.3 & 63.3 & 54.3 \\
  \midrule
  FastV & 68\% & 15.1 & 5.9 & 5.7 & 5.1 & 6.1 & 6.7 & 54.8 & 31.8 & 58.4 & 72.7 & 56.3 & 64.0 & 83.2 & 57.1 & 63.3 & 52.4\\
  PyramidDrop & 65\% & 15.7 & 28.9 & 24.0 & 29.2 & 29.7 & 32.9 & 60.8 & 47.1 & 61.2 & 76.9 & \bftable 57.9 & 65.3 & 86.6 & 58.2 & \bftable 63.4 & 53.1 \\
  \feather & 64\% & 15.7 & \bftable 39.3 & \bftable 33.1 & \bftable 40.1 & \bftable 39.7 & \bftable 44.1 & \bftable 61.9 & \bftable 51.4 & \bftable 61.8 & \bftable 77.9 & 56.5 & \bftable 66.1 & \bftable 87.7 & \bftable 59.1 & \bftable 63.4 & \bftable 54.2 \\
  \midrule
  FastV & 45\% & 16.8 & 29.1 & 17.5 & 29.5 & 33.1 & 36.1 & 61.0 & 45.8 & 62.3 & 77.4 & 58.4 & 65.7 & 86.8 & 59.2 & 63.3 & 53.5 \\
  PyramidDrop & 46\% & 16.8 & 46.6 & 37.4 & 48.3 & 47.8 & 53.0 & 63.7 & 53.8 & 63.1 & 78.7 & \bftable 59.1 & 66.2 & 87.5 & \bftable 59.4 & 63.5 & 54.3 \\
  \feather & 48\% & 16.5 & \bftable 49.7 & \bftable 39.3 & \bftable 52.1 & \bftable 50.9 & \bftable 56.7 & \bftable 63.9 & \bftable 54.6 & \bftable 63.2 & \bftable 78.8 & 59.0 & \bftable 66.3 & \bftable 87.7 & 59.2 & \bftable 64.0 & \bftable 54.6 \\
  \end{tabular}
\end{small}
\caption{Comparing \feather performance against FastV and PyramidDrop. The best results are \textbf{bolded} (excluding the baseline method).}
  \label{tab:feather_table}
  \end{minipage}
\end{table}

\begin{figure}[h]
  \centering
   \includegraphics[width=1\linewidth]{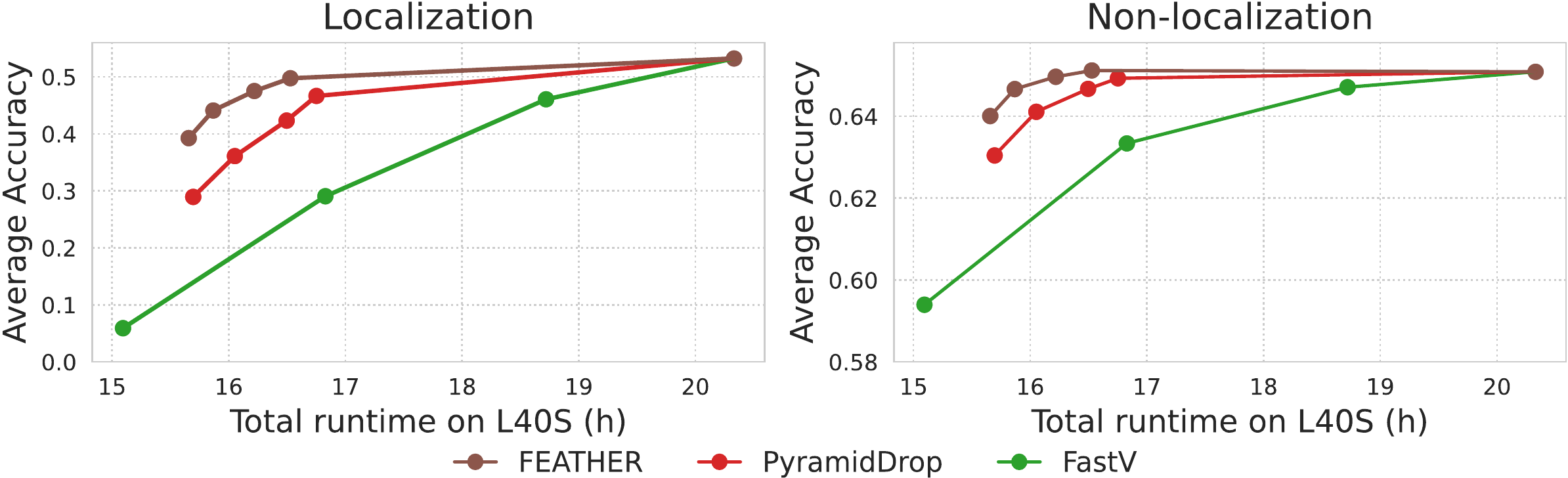}
   \caption{Total runtime on L40S vs. performance for FastV, PyramidDrop, and \feather.}
   \label{fig:time_results}
\end{figure}

\newpage
In addition, we show performance with respect to total runtime on a NVIDIA L40S in Figure~\ref{fig:time_results}.
\clearpage

\clearpage

\section{Comparison Against FasterVLM and\\VisionZip}

We present FasterVLM and VisionZip performance in Table~\ref{tab:additional_baselines_table}. We find that these approaches, while performing comparably to our approach on some benchmarks, perform vastly worse on localization benchmarks. We expect this is because positional information is not maintained in these methods, as image tokens are filtered without altering the positional embeddings.  We verify the importance of positional embeddings in \S\ref{sec:pos_importance}. Note that since our setup uses the SigLIP encoder, for FasterVLM (which relies on [CLS] attention), we use the proposed solution in VisionZip of averaging attention each token receives from all others in the sequence.

\section{Token Shuffling Ablation}
\label{sec:pos_importance}
To assess the impact of positional embeddings on model performance, we shuffle positional embeddings for the image tokens and evaluate both the original VLM and our \feather approach. As shown in Table~\ref{tab:additional_baselines_table}, the localization performance of both methods drops drastically for localization tasks, substantially for TextVQA, and relatively little for other benchmarks. This result supports our key insight that many vision-language benchmarks inadequately capture the shortcomings of efficiency methods due to their limited ability to assess fine-grained visual capabilities, particularly for visual grounding.

\vspace{2em}
\begin{table}[h]
  \centering
  \begin{minipage}{\textwidth}
    \centering
 \begin{small}
  \begin{tabular}{r@{\colpad}c@{\colpad}|c@{\colpad}c@{\colpad}c@{\colpad}c@{\colpad}c@{\colpad}|@{\colpad}c@{\colpad}c@{\colpad}c@{\colpad}c@{\colpad}c@{\colpad}|@{\colpad}c@{\colpad}c@{\colpad}c@{\colpad}c@{\colpad}c@{\colpad}}
  \multicolumn{2}{c}{} & \multicolumn{5}{c}{Localization} & \multicolumn{5}{c}{Open-Ended VQA} & \multicolumn{5}{c}{Challenge Sets} \\
   Method & \rotatecol{FLOPS Red \vphantom{X}} & \rotatecol{Avg} & \rotatecol{OCID-Ref} & \rotatecol{RefCOCOg} & \rotatecol{RefCOCO+} & \rotatecol{RefCOCO} & \rotatecol{Avg} & \rotatecol{TextVQA} & \rotatecol{GQA} & \rotatecol{VQAv2} & \rotatecol{VizWiz} & \rotatecol{Avg}  & \rotatecol{POPE} & \rotatecol{TallyQA} & \rotatecol{VSR} & \rotatecol{AI2D} \\
  \midrule
    Baseline &   0\% &  53.2 & 40.7 &  56.3 & 55.0 &  60.9 & 64.1 & 54.9 & 63.3 &  78.9 & 59.3 & 66.1 & 87.4 & 59.3 &  63.3 & 54.3 \\
   Baseline (pos shuffled) & 0\% & 8.0 & 9.0 & 7.8 & 7.1 & 8.0 & 59.2 & 44.1 & 60.3 & 75.8 & 56.8 & 63.3 & 86.6 & 55.3 & 59.2 & 51.9 \\
    \midrule
    FasterVLM & 65\% & 5.7 & 8.0 & 5.9 & 4.2 & 4.7 & 60.9 & 50.9 & 59.9 & 76.5 & 56.4 & \bftable 66.6 & 85.2 & 62.6 & \bftable 63.7 & \bftable 54.7 \\
    VisionZip & 65\% & 8.5 & 7.3 & 9.0 & 8.1 & 9.5 & 61.1 & 50.8 & 60.2 & 76.7 & \bftable 56.7 & 66.5 & 85.3 & \bftable 62.9 & \bftable 63.7 & 54.3\\
    \midrule
    \feather & 64\% & \bftable 39.3 & \bftable 33.1 & \bftable 40.1 & \bftable 39.7 & \bftable 44.1 & \bftable 61.9 & \bftable 51.4 & \bftable 61.8 & \bftable 77.9 & 56.5 & 66.1 & \bftable 87.7 & 59.1 & 63.4 & 54.2 \\
    \feather (pos shuffled) & 64\% & 5.3 & 5.3 & 4.8 & 5.2 & 5.8 & 57.8 & 41.7 & 58.9 & 75.0 & 55.5 & 63.2 & 86.0 & 55.7 & 58.8 & 52.5 \\
  \end{tabular}
\end{small}
\caption{Comparison against FasterVLM and VisionZip and positional embeddings ablation (where image token positions are shuffled). The best results are \textbf{bolded}.}
  \label{tab:additional_baselines_table}
  \end{minipage}
\end{table}

\newpage 
\section{Token Pruning Visualizations}
In this supplemental material section, we provide a qualitative analysis comparing the pruning effectiveness of various criteria as well as the final approaches of \feather, FastV, and PyramidDrop. Namely, we visualize the ability of approaches to retain important tokens, particularly for localization. In Figure~\ref{fig:k=3_criteria_visualization} and Figure~\ref{fig:k=8_criteria_visualization}, we visualize pruning from the various criteria assessed in the main text when pruning is done after layers three and eight, respectively. In Figure~\ref{fig:feather_comparison_visualization}, we visualize pruning from the final approaches of \feather, FastV, and PyramidDrop.


\subsection{Comparing pruning criteria}

\begin{figure*}[t]
    \centering
    \includegraphics[width=1\textwidth]{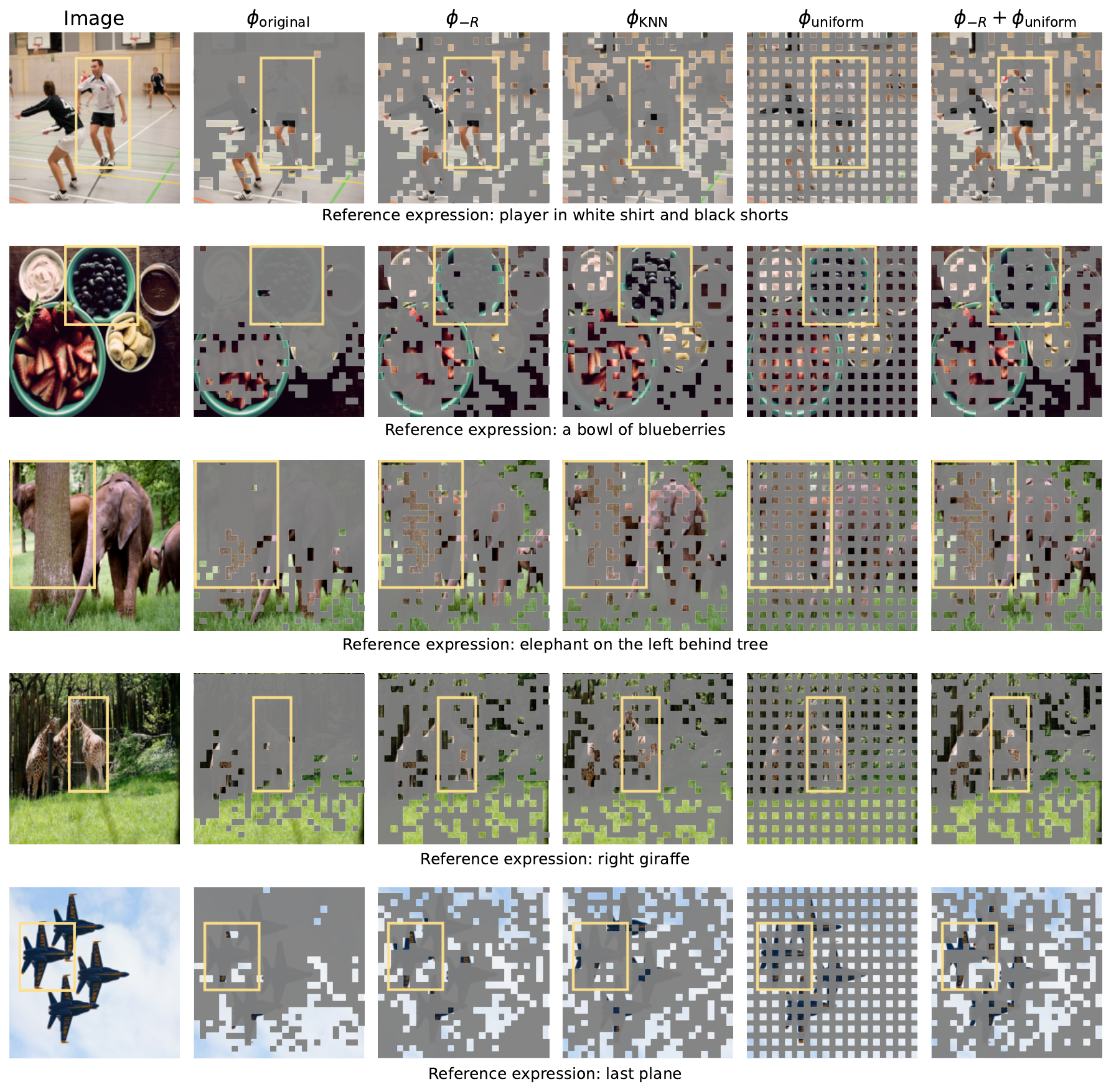}
    \captionof{figure}{Visualizing the ability of various pruning criteria to maintain visual tokens relevant to the reference expression when applied after layer three. We observe that $\phi_{\textbf{\textit{-R}}}$ resolves $\phi_{\text{original}}$'s tendency of selecting bottom image tokens and that uniform sampling is a robust approach that improves the token selection effectiveness of $\phi_{\textbf{\textit{-R}}}$ with $\phi_{\textbf{\textit{-R}}} + \phi_{\text{uniform}}$. See the main text for criteria definitions.}
    \label{fig:k=3_criteria_visualization}
\end{figure*}

\begin{figure*}[t]
  \centering
   \includegraphics[width=1\linewidth]{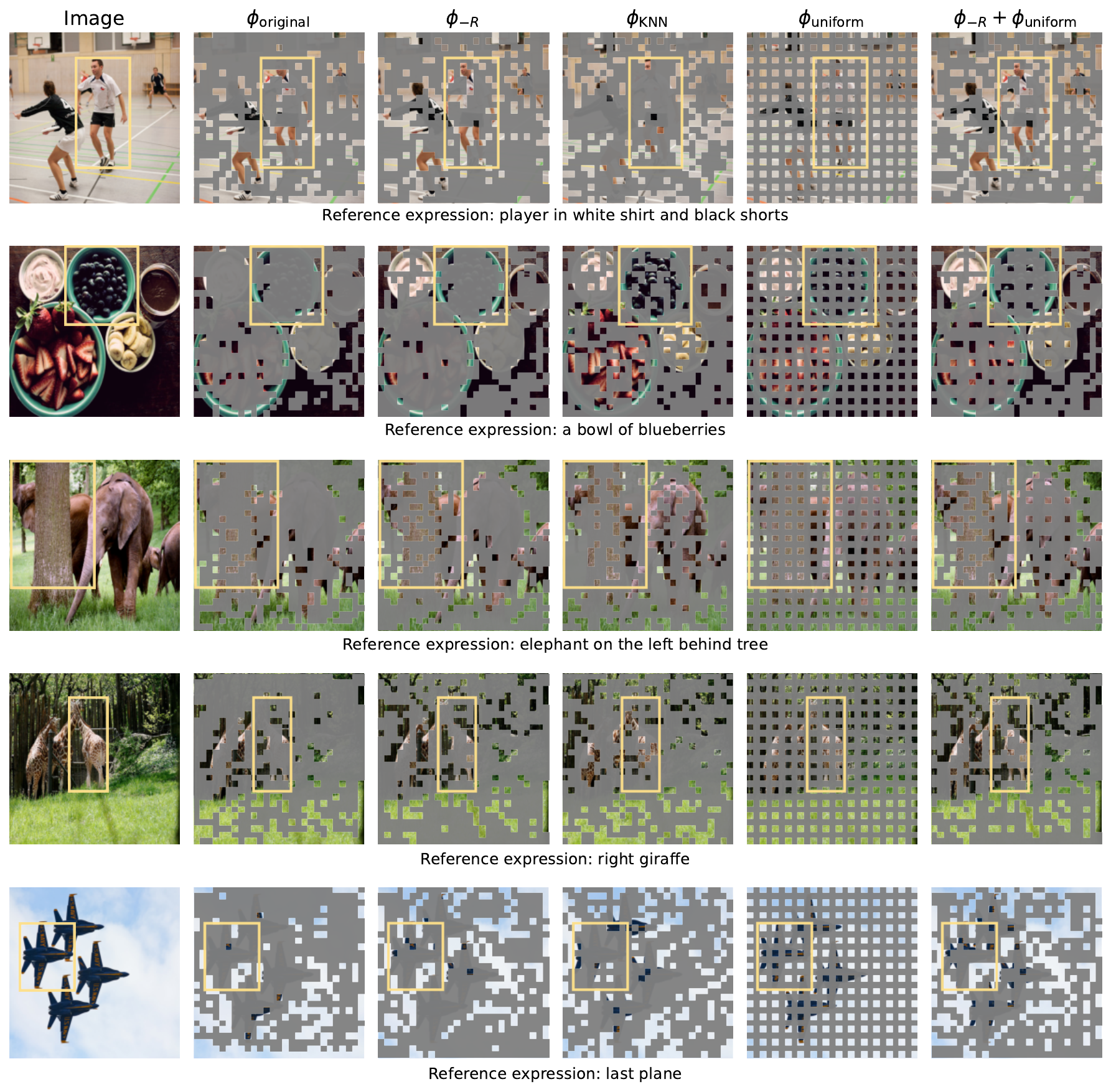}
   \caption{Visualizing the ability of various pruning criteria to maintain visual tokens relevant to the reference expression when applied after layer eight. We observe that the attention-based criteria are more effective when pruning after this layer compared to after layer three. See the main text for criteria definitions.}
   \label{fig:k=8_criteria_visualization}
\end{figure*}

We first visualize the retained tokens of various criteria when pruning is applied after layer three (see Figure~\ref{fig:k=3_criteria_visualization}) and layer eight (see Figure~\ref{fig:k=8_criteria_visualization}). We see that these visualizations support our quantitative results from the main paper. Specifically, (1) $\phi_{\textbf{\textit{-R}}}$ removes the criteria tendency of selecting bottom image tokens, resulting in an improved selection of maintained tokens; (2) the attention-based criteria improve when pruning after a later layer; and (3) adding uniform sampling to the attention-based pruning criteria with $\phi_{\textbf{\textit{-R}}} + \phi_{\text{uniform}}$ improves token selection.  

\clearpage

\clearpage

\clearpage

\subsection{Comparing \feather to FastV and PyramidDrop}

Additionally, we visualize the retained tokens for the \feather, FastV, and PyramidDrop approaches.

\newpage

As shown in Figure~\ref{fig:feather_comparison_visualization}, when comparing the remaining tokens used for prediction (after layer 16 for \feather, layer 24 for PyramidDrop, and layer three for FastV), we see that our approach retains substantially more tokens around and inside the reference expression bounding box.

\vspace{2em}
\noindent
\hspace*{-1.1\linewidth}
\begin{minipage}{\textwidth}
    \centering
    \includegraphics[width=1\textwidth]{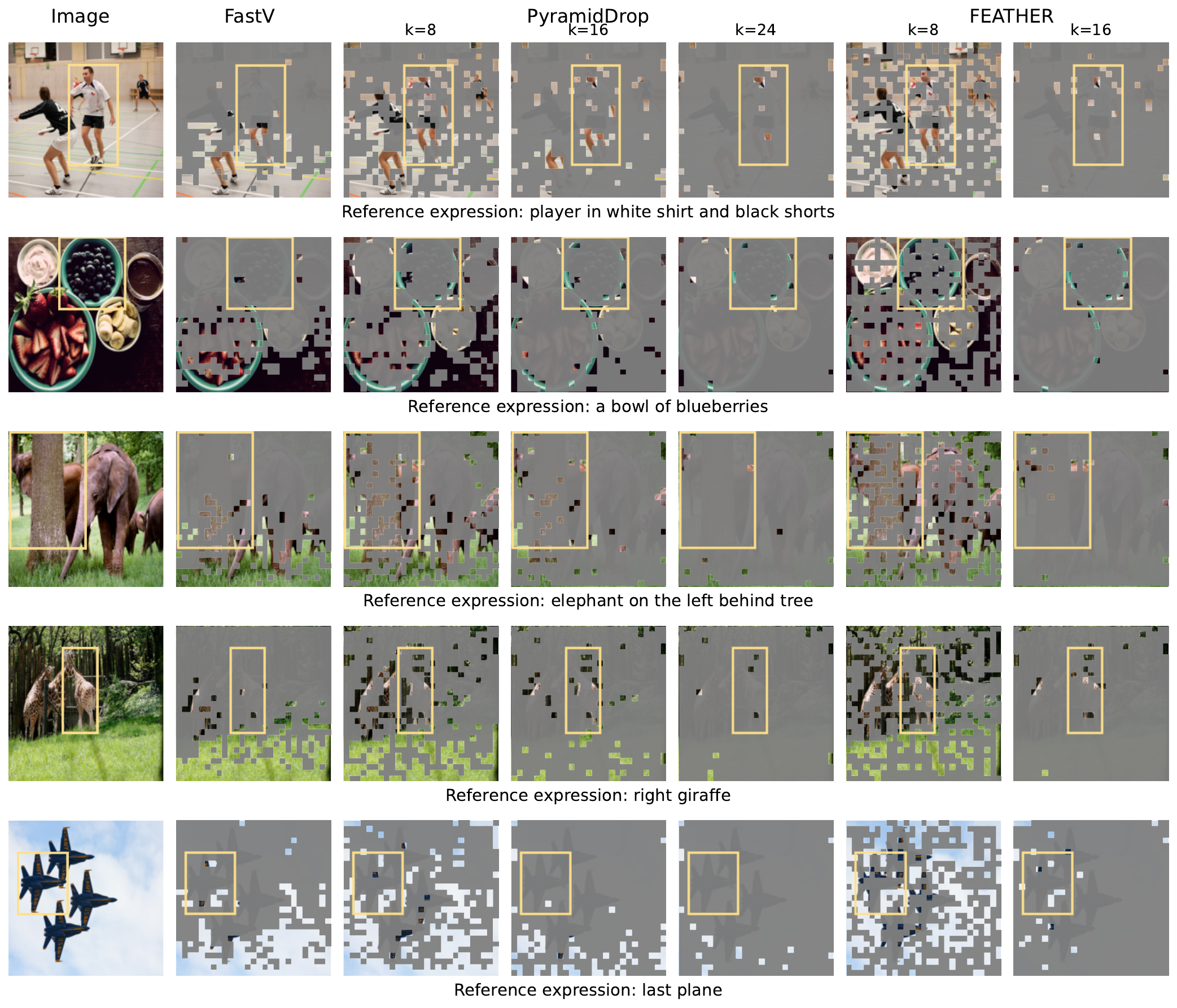}
    \captionof{figure}{Visualizing the ability of \feather, FastV, and PyramidDrop to retain visual tokens relevant to the reference expression. We observe that our approach retains a substantially higher portion of tokens relevant to the reference expression.}
    \label{fig:feather_comparison_visualization}
\end{minipage}

\end{document}